%% file: arxiv.tex
\documentclass[10pt,twocolumn,letterpaper]{article}

\usepackage[pagenumbers]{cvpr}

\usepackage{graphicx}
\usepackage{xcolor}
\usepackage{colortbl}
\usepackage{multirow}
\usepackage{algorithmic}
\usepackage{algorithm}
\usepackage{amsmath}
\usepackage{pgfplots}
\usepackage{subcaption}
\usepackage{placeins}

\usepackage{graphicx}
\usepackage{booktabs}
\usepackage[accsupp]{axessibility} 
\usepackage{hyperref}
\usepackage{orcidlink}
\usepackage{adjustbox}  
\usepackage{makecell}
\usepackage{multirow}
\usepackage{booktabs}

\definecolor{tabthird}{HTML}{ffffcc}
\definecolor{tabsecond}{HTML}{d9f0a3}
\definecolor{tabfirst}{HTML}{78c679}

\definecolor{colorTabTop}{rgb}{0.95,0.93,0.9} %
\definecolor{colorTab}{rgb}{0.9,0.97,0.9} %
\definecolor{color3}{rgb}{0.95,0.95,0.95}

\title{ChopGrad: Pixel-Wise Losses for Latent Video Diffusion \\via Truncated Backpropagation}

\author{
Dmitriy Rivkin$^{1}$,
Parker Ewen$^{1}$,
Lili Gao$^{1}$,
Julian Ost$^{1,2}$,
Stefanie Walz$^{1}$,\\
Rasika Kangutkar$^{1}$,
Mario Bijelic$^{1,2}$,
Felix Heide$^{1,2}$\\[6pt]
$^1$Torc Robotics, $^2$Princeton University\\
{\normalsize \textbf{}
}
}

\begin{document}
 \maketitle

\input{sections/00_Abstract} \label{sec:abs}
\input{sections/01_Introduction} \label{sec:intro}
\input{sections/02_Related_Work} \label{sec:rel_work}
\input{sections/03_Methodology} \label{sec:method}
\input{sections/04_Experiments} \label{sec:exp}
\input{sections/05_Conclusion} \label{sec:conc}
\section*{Acknowledgements}
Felix Heide was supported by an NSF CAREER Award (2047359), a Packard Foundation Fellowship, a Sloan Research Fellowship, a Sony Young Faculty Award, a Project X Innovation Award and a Amazon Science Research Award. Felix Heide is a co-founder of Algolux (now Torc Robotics), Head of AI at Torc Robotics, and a cofounder of Cephia AI.

\FloatBarrier

{
    \small
    \bibliographystyle{splncs04}
    \bibliography{main}
}
\clearpage
\appendix
\renewcommand{\theHsection}{appendix.\Alph{section}}
\renewcommand{\theHsubsection}{appendix.\Alph{section}.\arabic{subsection}}
\renewcommand{\theHsubsubsection}{appendix.\Alph{section}.\arabic{subsection}.\arabic{subsubsection}}
\section*{Appendix}
Section \ref{app:A} provides additional implementation information for the proposed \textit{ChopGrad} architecture using the WAN 2.1 \cite{wan2025wan} and CogVideoX \cite{yang2024cogvideox} video autoencoders. Next, Section \ref{app:B} reports additional details regarding evaluation setups and baseline implementations for all applications, while Section \ref{app:C} provides details about model architectures, training setups, and inference schemes.
Next, Section \ref{app:D} describes additional algorithmic optimizations for \textit{ChopGrad} to minimize computation times when long truncation distances are used.
Finally, Sections \ref{app:E} and \ref{app:F} report additional quantitative and qualitative results, respectively.

\input{sections/06_Appendix_A}\label{app:A}

\input{sections/07_Appendix_B}
\input{sections/08_Appendix_C}
\input{sections/09_Appendix_D} \label{app:D}
\input{sections/11_Appendix_F} \label{app:F} 
\input{sections/10_Appendix_E} \label{app:E} 

\begin{figure*}[!t]
    \centering
    \begin{minipage}{0.7\linewidth}
          \begin{minipage}{0.32\linewidth}\centering High-Res\end{minipage}
        \begin{minipage}{0.2\linewidth}\centering Low-Res\end{minipage}
        \begin{minipage}{0.2\linewidth}\centering DOVE\end{minipage}
        \begin{minipage}{0.2\linewidth}\centering ChopGrad\end{minipage}
        \includegraphics[width=\linewidth]{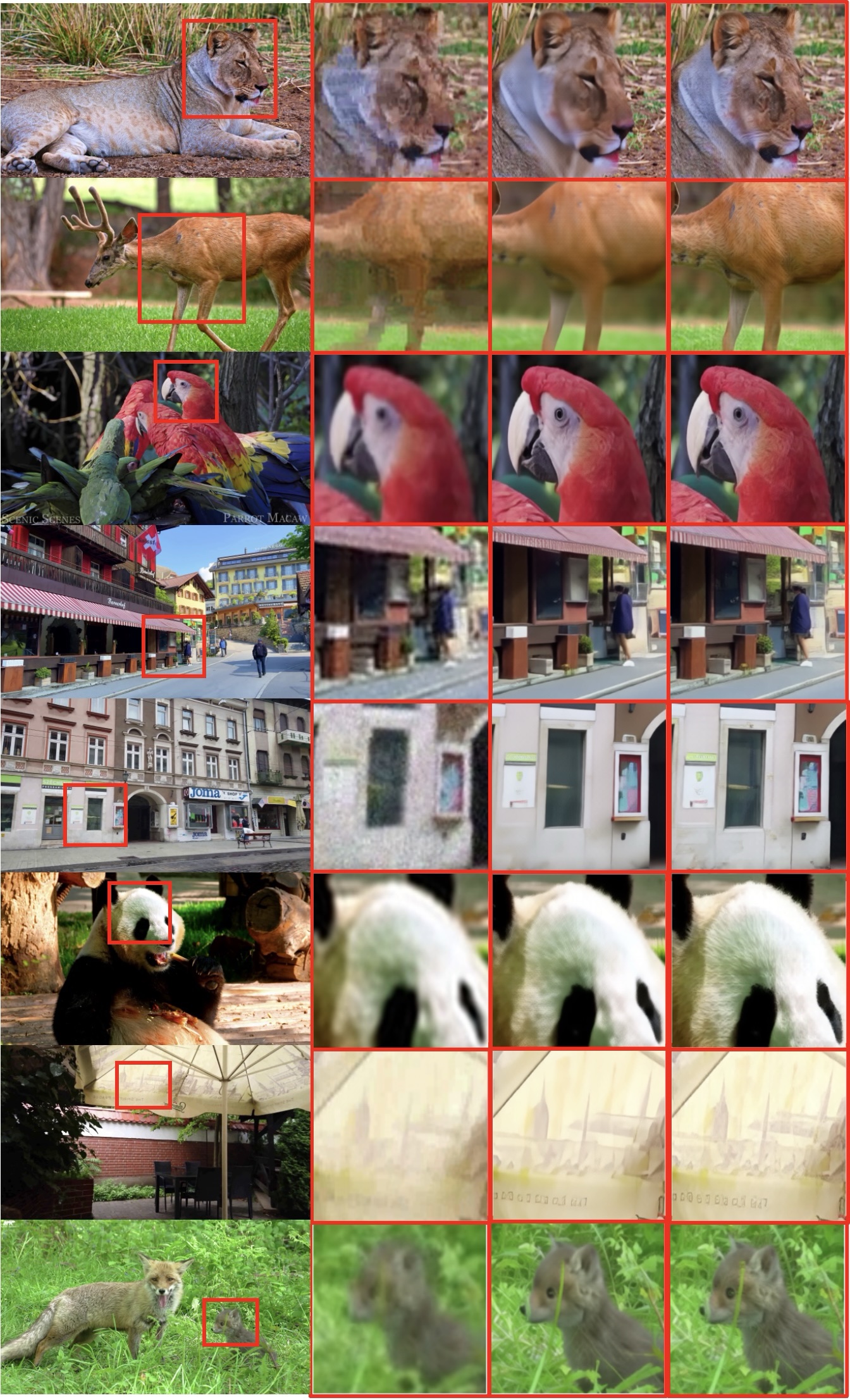}
    \end{minipage}
        \caption{Additional Video Super-Resolution Comparison. Shown from left to right: high-resolution, low-resolution input, DOVE \cite{chen2025dove}, and the proposed approach, \textit{ChopGrad}. \textit{ChopGrad} synthesizes fine textures better and reduces motion blur, especially in regions with high-frequency details like fur, hair, cloth, and clouds.}
    \label{fig:super_res_appendix}
\end{figure*}

\begin{figure*}[!t]
    \centering
    \begin{minipage}{0.95\linewidth}
        \begin{minipage}{0.16\linewidth}\centering Ground Truth\end{minipage}
        \begin{minipage}{0.16\linewidth}\centering 3DGS\end{minipage}
        \begin{minipage}{0.16\linewidth}\centering MVSplat-360\end{minipage}
        \begin{minipage}{0.16\linewidth}\centering Difix\end{minipage}
        \begin{minipage}{0.16\linewidth}\centering ChopGrad $D_{trunc}=1$\end{minipage}
        \begin{minipage}{0.16\linewidth}\centering ChopGrad $D_{trunc}=2$\end{minipage}
    \includegraphics[width=\linewidth]{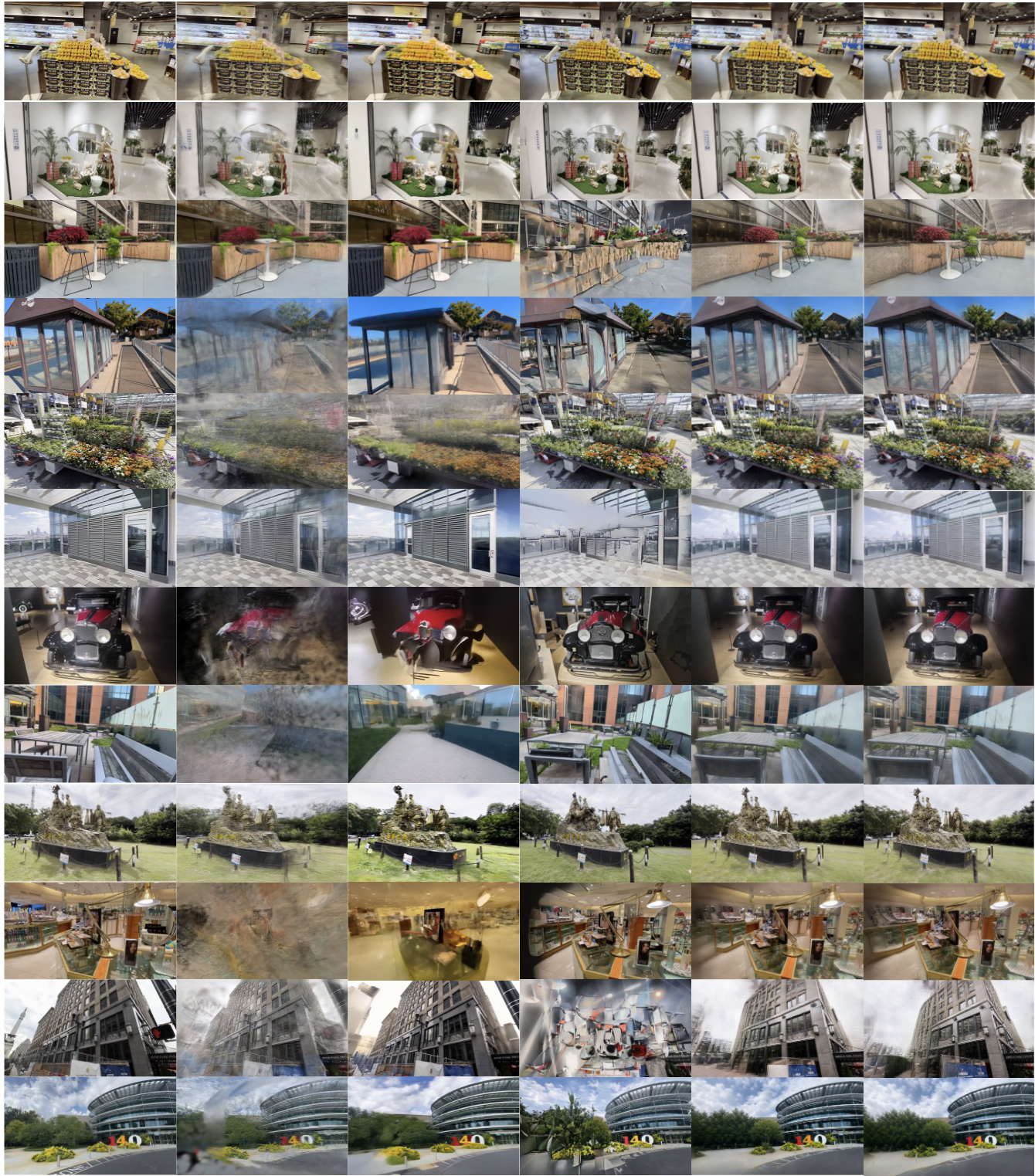}
    \end{minipage}
    \caption{Additional Qualitative Results for Artifact Removal in Novel View Synthesis on the DL3DV-Benchmark Dataset \cite{ling2024dl3dv}. Ground truth video frames and 3DGS model renders are shown on the left.  Results for MVSplat-360 \cite{chen2024mvsplat360} and Difix \cite{wu2025difix3d} are presented alongside the \textit{ChopGrad} with a truncation distance of $1$ and $2$. \textit{ChopGrad} corrects significantly more artifacts than other methods (e.g., fourth row from the top) with fewer hallucinations (e.g., 5th row from the bottom), and maintains temporal consistency over the entire video sequence.}
    \label{fig:3dgs_appendix}
\end{figure*}

\begin{figure*}[!t]
    \centering
    \begin{minipage}{0.69\linewidth}
    \begin{minipage}{0.32\textwidth}\centering VACE\end{minipage}
    \begin{minipage}{0.32\textwidth}\centering ChopGrad\end{minipage}
    \begin{minipage}{0.32\textwidth}\centering Ground Truth\end{minipage}
    \includegraphics[width=\linewidth]{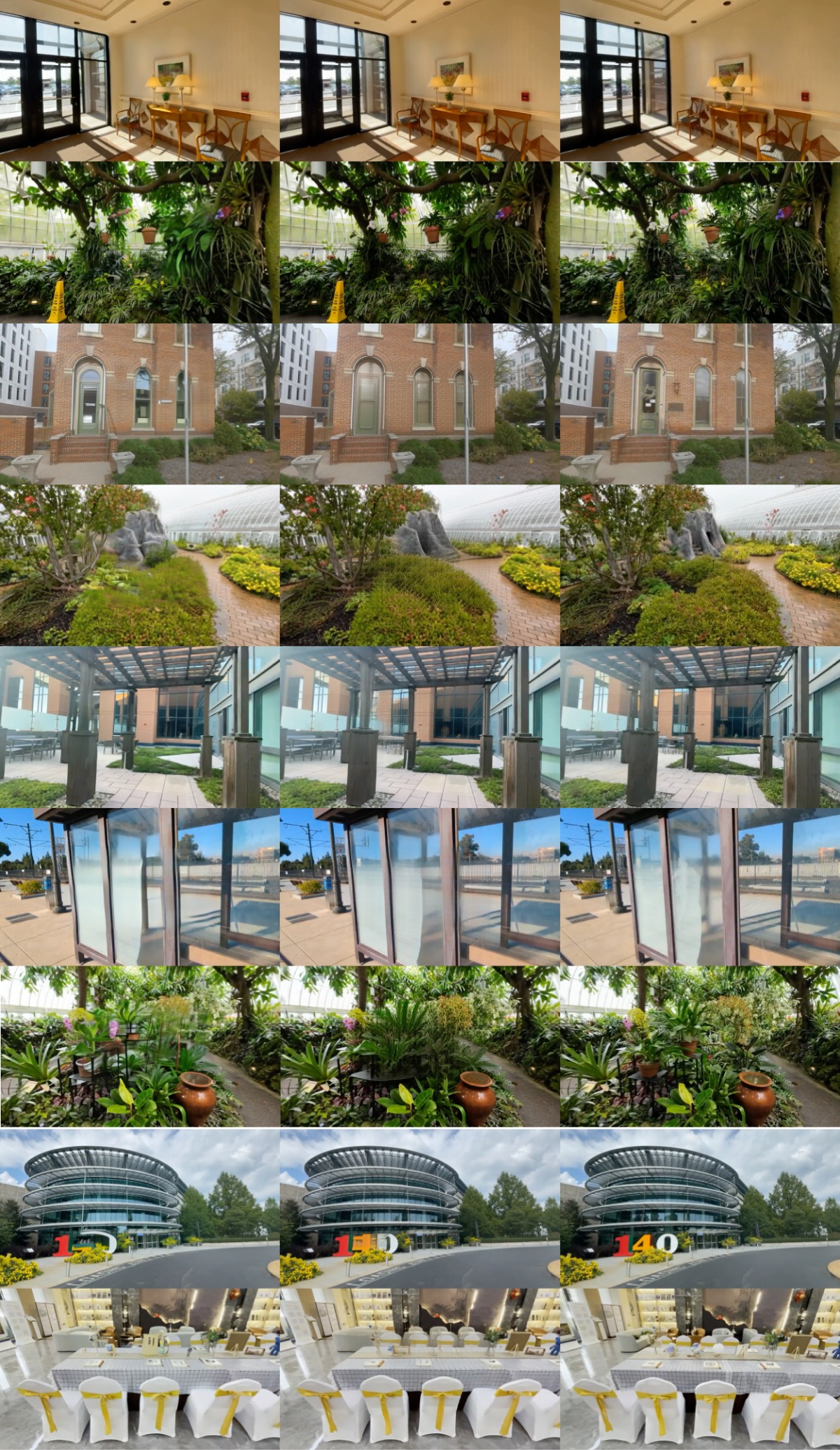} 
    \end{minipage}
    \caption{Additional Video Inpainting Comparison on D3LDV Dataset. Shown from left to right: VACE, \textit{ChopGrad}, Ground Truth. Training with \textit{ChopGrad} reduces hallucinations despite $50\times$ lower inference budget.}
    \label{fig:inpainting_appendix_d3ldv}
\end{figure*}

\begin{figure*}[!t]
    \centering
    \begin{minipage}{0.69\linewidth}
    \begin{minipage}{0.32\linewidth}\centering VACE\end{minipage}
    \begin{minipage}{0.32\linewidth}\centering ChopGrad\end{minipage}
    \begin{minipage}{0.32\linewidth}\centering Ground Truth\end{minipage}
    \includegraphics[width=\linewidth]{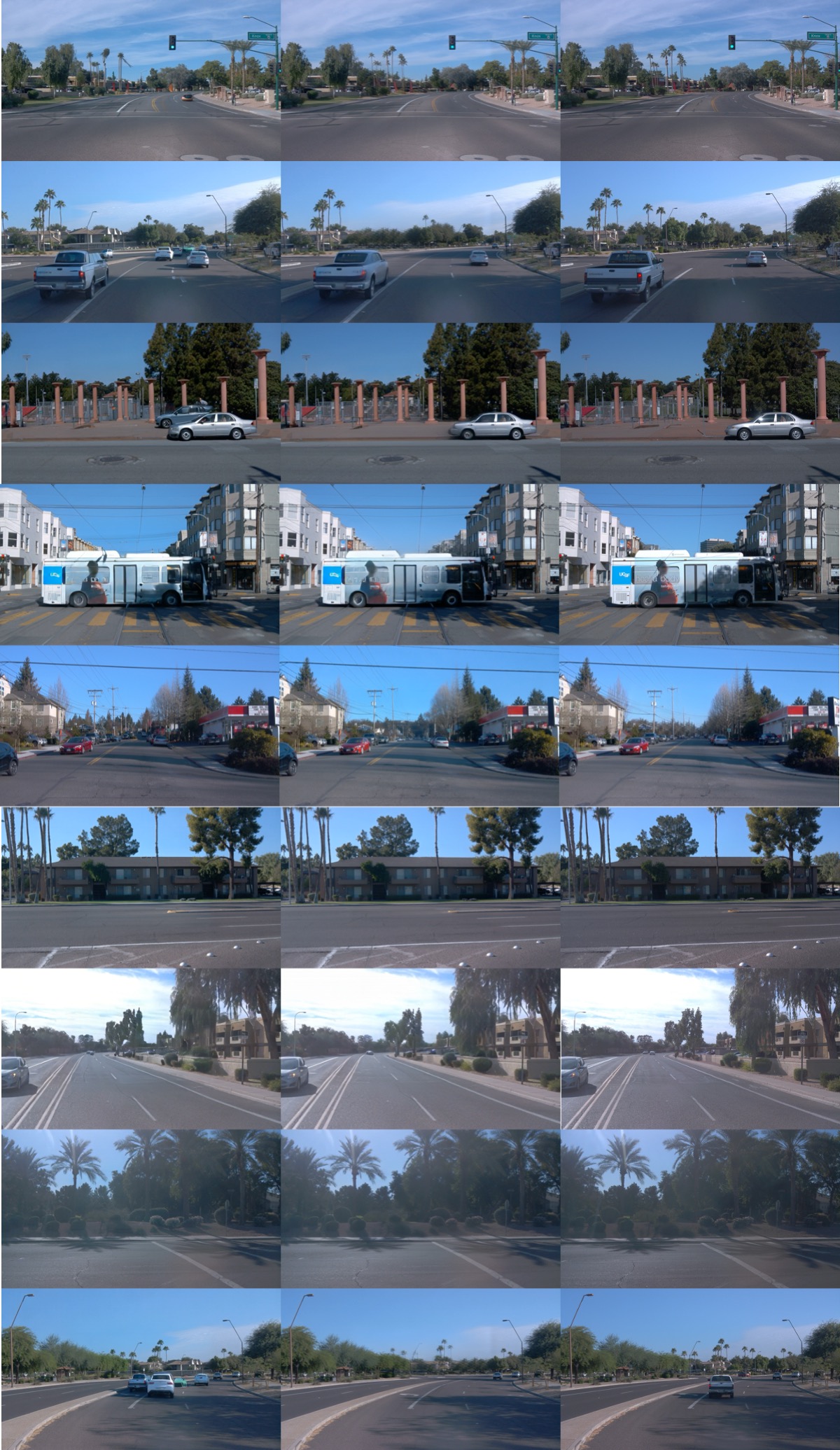}
    \end{minipage}
    \caption{Additional Video Inpainting Comparison on Waymo Dataset.  Shown from left to right: VACE, \textit{ChopGrad}, Ground Truth. Training with \textit{ChopGrad} reduces hallucinations despite $50\times$ lower inference budget.
    }
    \label{fig:inpainting_appendix_waymo}
    
\end{figure*}
\begin{figure*}[!t]
    \centering
    \begin{minipage}{0.69\linewidth}
    \begin{minipage}{0.24\linewidth}\centering VACE\end{minipage}
    \begin{minipage}{0.24\linewidth}\centering ChopGrad\end{minipage}
    \begin{minipage}{0.24\linewidth}\centering Ground Truth\end{minipage}
    \begin{minipage}{0.24\linewidth}\centering Mask\end{minipage}
    \includegraphics[width=\linewidth]{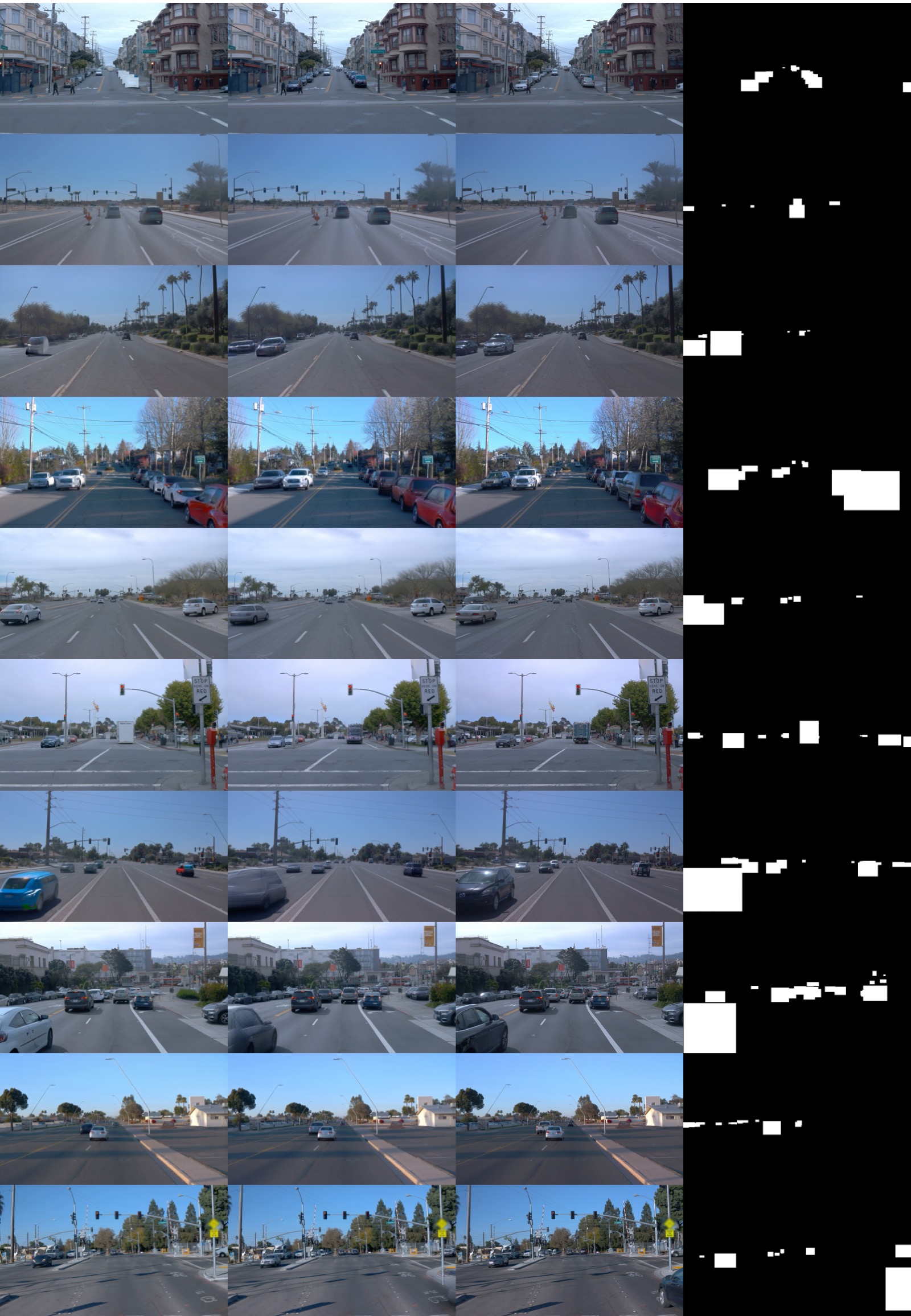}
    \end{minipage}
        \caption{Additional Video Inpainting Comparison on Waymo-Bbox Task. In this task, 50\% of the vehicles are randomly selected for masking. Shown from left to right: VACE, \textit{ChopGrad}, Ground Truth. Training with \textit{ChopGrad} reduces hallucinations despite $50\times$ lower inference budget.
        }
    \label{fig:inpainting_appendix_waymo_bbox}
\end{figure*}

\begin{figure*}[!t]
    \centering
    \begin{minipage}{0.7\linewidth}
    \begin{minipage}{0.24\linewidth}\centering VACE\end{minipage}
    \begin{minipage}{0.24\linewidth}\centering ChopGrad\end{minipage}
    \begin{minipage}{0.24\linewidth}\centering Ground Truth\end{minipage}
    \begin{minipage}{0.24\linewidth}\centering Mask\end{minipage}
    \includegraphics[width=\linewidth]{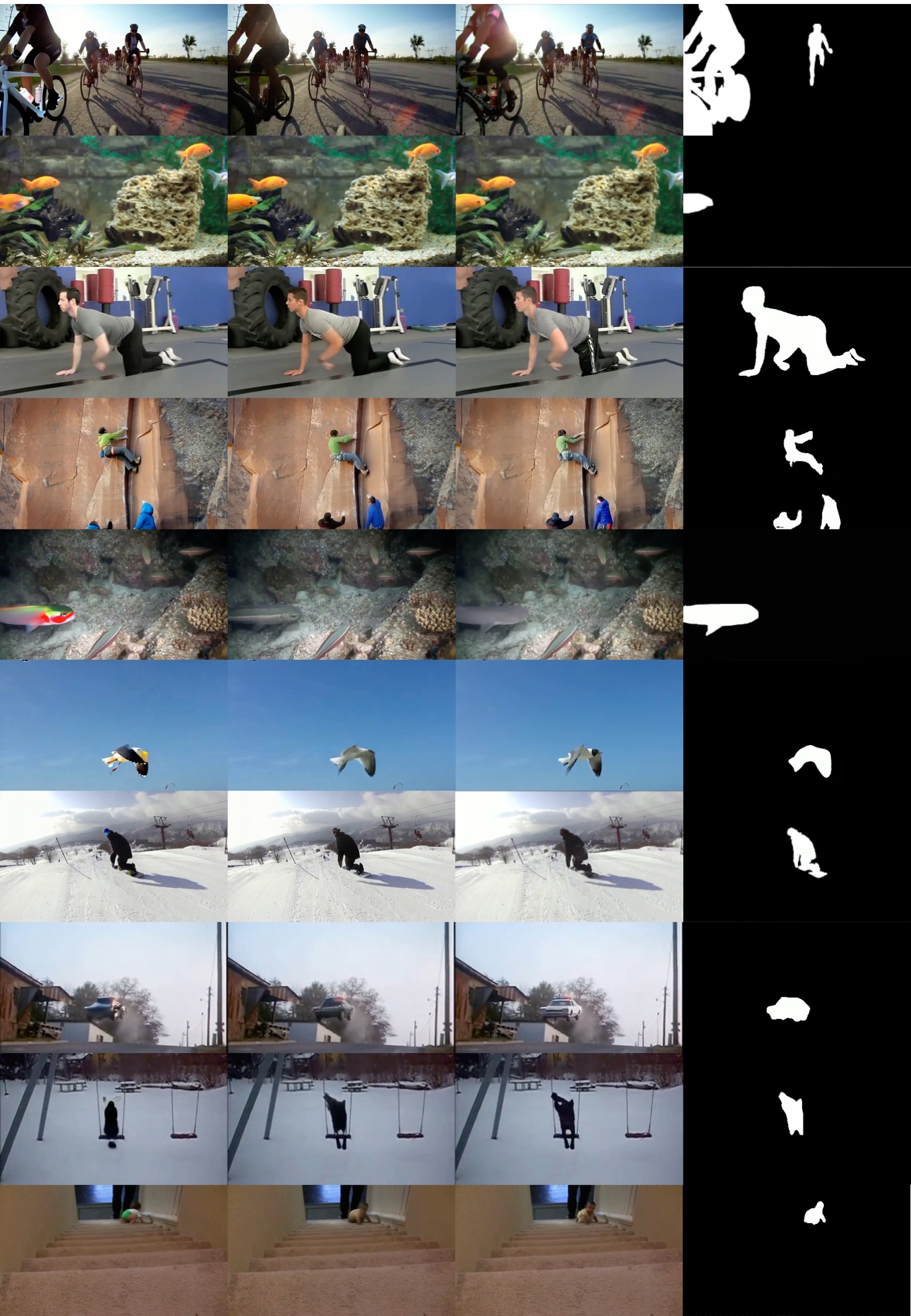}
    \end{minipage}
        \caption{Additional Video Inpainting Comparison on ROVI Dataset. Shown from left to right: VACE, \textit{ChopGrad}, Ground Truth. Training with \textit{ChopGrad} reduces hallucinations despite $50\times$ lower inference budget.
        }
    \label{fig:inpainting_appendix_rovi}
\end{figure*}

\begin{figure*}[!t]
    \centering
    \begin{minipage}{0.7\linewidth}
    \begin{minipage}{0.24\linewidth}\centering Naive Insertion\end{minipage}
    \begin{minipage}{0.24\linewidth}\centering Mirage\end{minipage}
    \begin{minipage}{0.24\linewidth}\centering ChopGrad\end{minipage}
    \begin{minipage}{0.24\linewidth}\centering Ground Truth\end{minipage}
            \includegraphics[width=\linewidth]{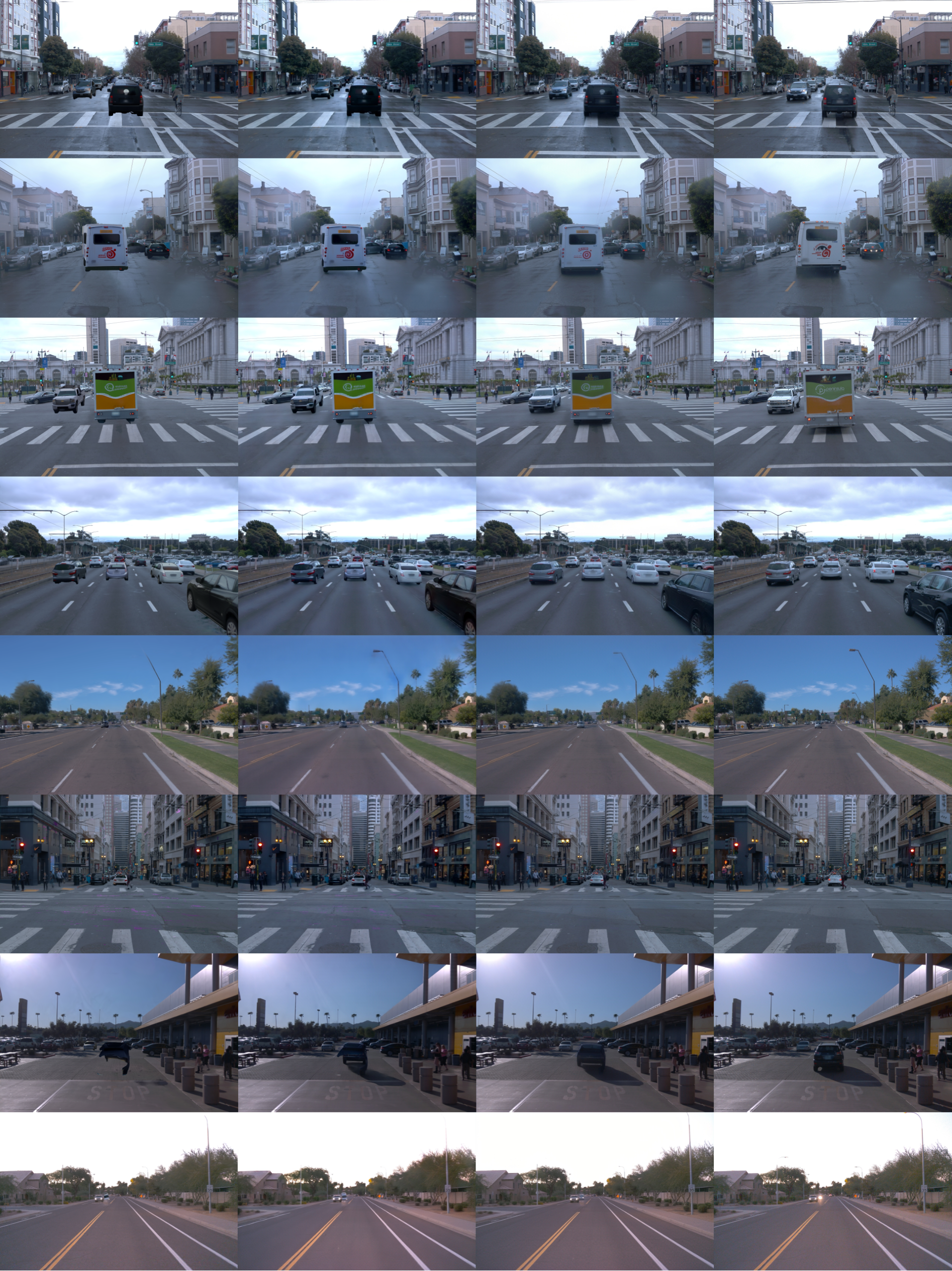}
    \end{minipage}
        \caption{Additional Controlled Driving Video Generation Comparison. Shown from left to right: Naive Insertion, Mirage \cite{wang2025mirage}, \textit{ChopGrad}, and Ground Truth. The top six rows demonstrate that training with \textit{ChopGrad} increases realism by improving lighting and shadows, and removing more Gaussian Splat artifacts. The bottom two rows, which have been cropped, demonstrate that training with \textit{ChopGrad} enables the model to make stronger collections in the presence of very poor vehicle model quality -- note that these very poor vehicle models are relatively rare in the dataset.}
    \label{fig:asset_variation_appendix}
\end{figure*}

\end{document}

%% file: sections/00_Abstract.tex
\begin{abstract}
Recent video diffusion models achieve high-quality generation through recurrent frame processing where each frame generation depends on previous frames. However, this recurrent mechanism means that training such models in the pixel domain incurs prohibitive memory costs, as activations accumulate across the entire video sequence.
This fundamental limitation also makes fine-tuning these models with pixel-wise losses computationally intractable for long or high-resolution videos. This paper introduces ChopGrad, a truncated backpropagation scheme for video decoding,  limiting gradient computation to local frame windows while maintaining global consistency. We provide a theoretical analysis of this approximation and show that it enables efficient fine-tuning with frame-wise losses. ChopGrad reduces training memory from scaling linearly with the number of video frames (full backpropagation) to constant memory, and compares favorably to existing state-of-the-art video diffusion models across a suite of conditional video generation tasks with pixel-wise losses, including video super-resolution, video inpainting, video enhancement of neural-rendered scenes, and controlled driving video generation. Our project page is available at \url{https://light.princeton.edu/chopgrad}.
\end{abstract}

%% file: sections/01_Introduction.tex
\vspace{-12pt}\section{Introduction}
Recent methods in latent video diffusion are capable of generating high-resolution videos over long time horizons \cite{wan2025wan, wang2025lavie, he2022latent, xing2024survey}.
Similar to latent image diffusion models, latent video diffusion models rely on pre-trained autoencoders to compress videos into latent embeddings and then learn over these embeddings \cite{hacohen2024ltx, blattmann2023align, ceylan2023pix2video}.
An enabling factor for recent video diffusion results is the use of temporal compression, where the autoencoder not only compresses video frames along spatial dimensions, but also along the temporal dimension~\cite{melnik2024video, zhou2024allegro, yu2024viewcrafter}.

\begin{figure}[t]
    \centering
    \includegraphics[width=0.9\linewidth,clip]{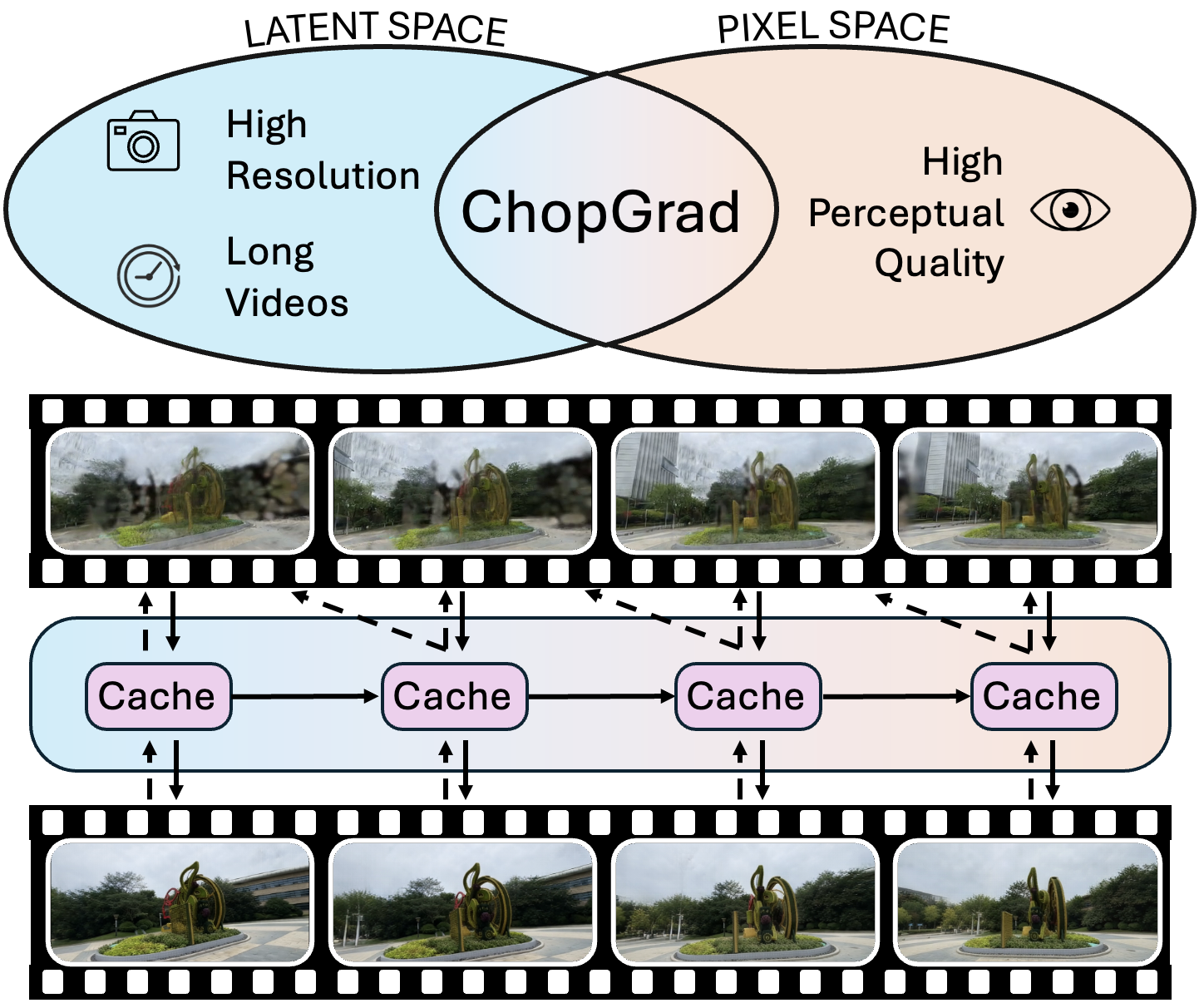}
    \caption{\textbf{ChopGrad Method}. 
    \textit{ChopGrad} unlocks pixel-wise losses for high resolution, long-duration video diffusion models. It leverages truncated backpropagation to eliminate recursive activation accumulation in video autoencoders with causal caching. Solid arrows indicate the flow of information in the decoder forward pass, dashed ones indicate the backward flow of gradients with \textit{ChopGrad}. Adding \textit{ChopGrad} to training procedures is easy and produces state of the art performance in a variety of applications that benefit from pixel-wise losses, such as video super-resolution, video inpainting, video enhancement of neural rendered scenes, and controlled driving video generation.
    }
    \label{fig:main} 
    \vspace{-0.5cm}
\end{figure}

Temporal compression groups multiple image frames into a single latent frame group. To incentivize temporal consistency between these frame groups causal caching has been introduced \cite{yang2024cogvideox, yu2023language, wu2025improved}.
This technique appends embeddings from previous frame group encodings onto the beginning of subsequent frame groups at each layer of the video encoder and decoder.
Notably, this approach introduces a recurrent structure into the autoencoder, where the dependency graph of video latents requires gradients to be propagated through all previous frame embeddings.

At the same time, most successful latent video diffusion models are trained within the latent space \cite{an2023latent, he2022latent, yang2024cogvideox, wan2025wan}, meaning gradients are not propagated through the encoder or decoder during latent video diffusion training. As such, existing methods make \emph{pixel-wise losses intractable} for long-duration videos as the gradients of these losses require the recurrent accumulation of activations through the decoder. These pixel-level perceptual losses are used extensively in finetuning image diffusion models and video models with \emph{short-duration}, low-resolution videos in applications such as single-step model distillation \cite{dmd}, enhancement of neural rendered scenes \cite{wu2025difix3d, chen2024mvsplat360}, image translation \cite{pix2pixturbo}, video super-resolution \cite{chen2025dove}, and controlled driving video generation \cite{ljungbergh2025r3d2, wang2025mirage}. 
In work such as \cite{pix2pixturbo, wu2025difix3d, ljungbergh2025r3d2, wang2025mirage}, the decoder itself is finetuned, making support for pixel-wise losses a strict requirement for training these types of models.

To enable pixel-wise losses for high-resolution, long duration video diffusion, this work introduces \textit{ChopGrad}, a truncated backpropagation scheme for video decoding (Fig. \ref{fig:main}).
Truncated backpropagation prevents activation accumulation over the full unrolled network by limiting the number of previous frames the gradients can propagate through.
To validate this, we define latent temporal locality to demonstrate that the effect of prior video frames in the gradient error drops off at an exponential rate.
We show that the proposed method enables efficient training using pixel-wise losses, such as the LPIPS \cite{zhang2018unreasonable} loss, across a variety of tasks and multiple video diffusion models. We evaluate our method on several applications, including video super-resolution, video inpainting, video enhancement of neural rendered scenes, and controlled driving video generation, outperforming existing latent video diffusion adaptation methods in terms of quantitative frame-wise and video performance metrics. These results are achieved with modest computational resources (training times of approximately 3 to 4 hours on 4 to 8 A100 GPUs). The contributions of this paper are:
\begin{itemize}
    \item A mathematical derivation and error analysis of truncated backpropagation for causal video autoencoders,
    \item A memory-efficient, practical approach for implementing pixel-wise losses for fine-tuning latent video diffusion models that generalizes across multiple diffusion models,
    \item Validation of the method across several tasks requiring pixel-wise losses, including video super-resolution, video inpainting, video enhancement of neural-rendered scenes, and controlled driving video generation, comparing favorably to existing baselines in all experiments.
\end{itemize}

\begin{figure*}[t]
    \centering
    \includegraphics[width=0.95\linewidth,clip]{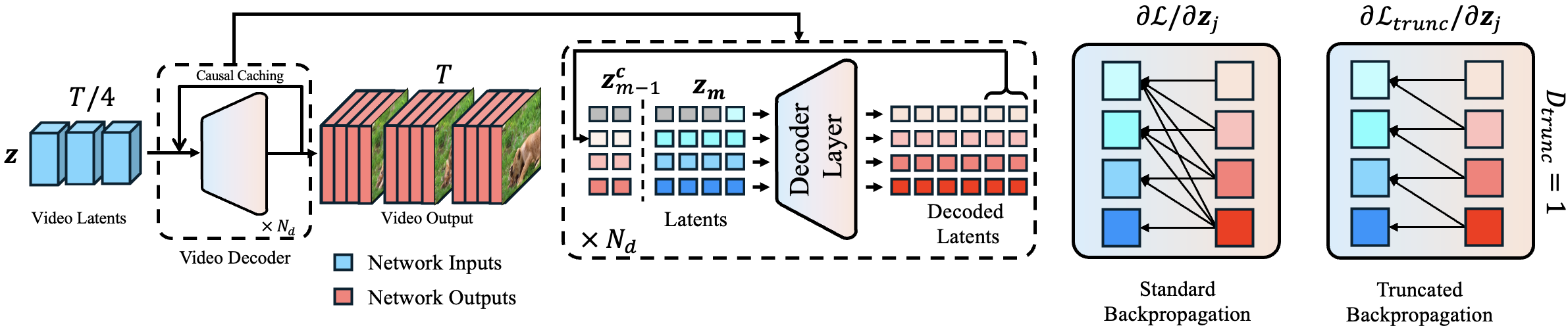}
    \caption{\textbf{ChopGrad Model Architecture}. Given the processed video frame latents, the video decoder iteratively applies causal caching at each layer, producing pixel outputs. Caching is performed by taking a subset of the layer outputs and appending these to the beginning of the layer inputs for the next frame group. While substantially reducing memory use at inference time compared to full 3D convolution over all frame groups, during training this process introduces recursive activation accumulation in the decoder, making backpropagation prohibitively expensive for high-resolution or long videos when using pixel-wise losses. Using truncated backpropagation, we only allow gradients to accumulate through a fixed number ($D_{trunc}$) of previous frame groups.}
    \label{fig:network_diagram} 
    \vspace{-0.5cm}
\end{figure*}

%% file: sections/02_Related_Work.tex
\section{Related Work}
Latent video diffusion has experienced rapid advancement in recent years thanks in part to novel video auto-encoding methods \cite{xing2024survey, ceylan2023pix2video, chen2024videocrafter2, melnik2024video}.
In particular, temporal compression and causal caching have demonstrated significant improvements in video quality and temporal consistency.

Latent video diffusion models extend latent image diffusion methods to model temporally coherent video sequences by operating in a compressed latent space rather than pixel space \cite{ho2022video, xing2024survey, blattmann2023align, yu2023video}. 
Operating in a latent space \cite{blattmann2023stable, wang2025lavie, singer2022make, li2024drivingdiffusion} reduces per-frame dimensionality and enables tractable scaling to longer and higher-resolution clips while preserving perceptual fidelity \cite{ho2022imagen, he2022latent}.
Early video diffusion formulations applied standard image-based diffusion techniques directly to short clips, jointly denoising fixed-length frame blocks and introducing conditioning strategies to extend temporal length \cite{ho2022video, singer2022make, an2023latent, danier2024ldmvfi}.

One of the most prevalent architectural advancements powering latent video diffusion is the use of temporal compression \cite{golinski2020feedback, d2017autoencoder, hacohen2024ltx, zhou2024allegro, zheng2024open, chen2024od} and causal caching to preserve latent integrity and temporal consistency when processing long sequences \cite{yu2023language, an2023latent, gao2024ca2}.
Causal caching has been used to maintain reconstruction fidelity and avoid temporal flicker while dramatically reducing memory and latency during encoding/decoding \cite{li2025wf, wu2025improved}. 
Unfortunately, this causal caching mechanism for video encoding introduces a recurrent structure into the encoders and decoders used by latent video diffusion models, resulting in prohibitive memory consumption due to activation accumulation during training when pixel-wise losses are used. 

A similar problem was encountered in early natural language processing with recurrent neural networks \cite{rumelhart1985learning, salehinejad2017recent, pascanu2013difficulty}, where truncated backpropagation through time was used to mitigate this issue \cite{williams2013gradient,aicher2020}. To the best of our knowledge, this paradigm has not been investigated or applied for image or video models.

Diffusion models often require long inference times, as the model must be run many times to generate an output. Single and few-step distillation \cite{dmd, sdturbo, self-forcing, yang2025towards, wang2025videoscene, noroozi2024you, chadebec2025flash, mao2025osv} has been used to reduce the number of steps required. Single-step distillation has also been used to adapt diffusion models to image-to-image translation tasks like changing weather or generating images from sketches \cite{pix2pixturbo,lee2025single,chen2025genhaze}. In applications where input/output pairs are readily available (such as super-resolution \cite{chen2025dove, he2024one, wang2024sinsr, wang2025seedvr2} or 3D gaussian splatting post-processing \cite{wu2025difix3d,wang2025mirage,dong2026one, ljungbergh2025r3d2}), pretrained diffusion models \cite{chen2025dove, wang2025mirage, dong2026one, teng2025gfix} or their one-step distilled counterparts \cite{wu2025difix3d,ljungbergh2025r3d2} have been finetuned for single-step inference on the given task. Many of these single-step distillation and finetuning approaches rely on pixel-wise perceptual losses, albeit at low resolution and video duration in the case of video models due to memory constraints. As such, these single-step diffusion applications can derive the most benefit from \textit{ChopGrad}.

\subsection{Preliminaries}

Latent video diffusion models work by first mapping from the high-dimensional pixel space to a lower-dimensional latent space, down-sampling both the spatial and temporal dimensions via a pre-trained 3D VAE video encoder \cite{yang2024cogvideox, blattmann2023stable}.
Once encoded, the video embeddings are then processed by the network backbone, often a transformer, which learns the temporal evolution of the video embeddings.
Finally, the output embeddings are re-projected into pixel space via the pre-trained 3D VAE decoder.

The structure of such 3D VAE networks groups a set of frames into a single latent embedding.
To retain temporal consistency these networks use what is called causal logic padding \cite{wu2025improved} or causal caching \cite{li2025wf}, where the trailing $N$ outputs from the previous frame group are concatenated to the beginning of the subsequent frame group at each layer of the encoder and decoder \cite{yang2024cogvideox, yu2023language}.
This results in a recurrent structure, where the gradients of pixel-wise losses on later frames propagate through all previous frame groups.

When training 3D VAEs, computational resources are dedicated solely to the VAE, and approaches such as sequence parallelism can be used to mitigate these issues, as described in \cite{yang2024cogvideox}. 
In addition, 3D VAEs are also able to be trained at lower resolutions/durations with results generalizing to higher resolution/duration videos with no additional fine-tuning \cite{yang2024cogvideox}.
However, when training or fine-tuning latent video diffusion model transformers or U-nets, the majority of the memory budget is consumed by these backbones, prohibiting the allocation of significant memory resources to decoder backpropagation. 
The backbones must also be trained at high resolution/duration if they are to perform well for high-resolution/duration inference, further compounding these memory requirements, especially as adding pixel-wise losses also requires the decoders to perform inference at high resolution/duration, even if their own parameters are frozen.

%% file: sections/03_Methodology.tex
\section{ChopGrad}

In order to enable training of video diffusion models on long, high-resolution videos with pixel-wise losses while maintaining modest memory requirements we present \textit{ChopGrad}, a novel method for backpropagating through the video decoder.
Sections \ref{subsec:causal_caching} and \ref{subsec:trunc_back} report that popular pre-trained video autoencoders with causal caching demonstrate temporal locality, where frame groups only affect other frame groups in close temporal proximity.
Motivated by this insight, \textit{ChopGrad} applies truncated backpropagation through time to the decoder cache to increase computational efficiency with minimal degradation in performance.
With truncated backpropagation, gradients of each frame group are only able to accumulate to a portion of prior frame groups set by the truncation distance.
This breaks the recursive loop present in popular video autoencoders and enables pixel-wise losses for long, high-resolution videos.
In Section \ref{subsec:analysis} we quantify temporal locality and truncation gradient error in the Wan2.1 decoder and transformer. Implementation details are provided in the Appendix.

\subsection{Causal Caching in Temporal VAEs} \label{subsec:causal_caching}

The temporal VAE architecture with causal masking is first formalized. 
Let $\mathbf{X} = \{\mathbf{x}_1, \mathbf{x}_2, \ldots, \mathbf{x}_T\}$ denote a video sequence of $T$ frames, where each frame $\mathbf{x}_t \in \mathbb{R}^{H \times W \times C}$ has height $H$, width $W$, and $C$ channels.

The 3D VAE encoder groups consecutive frames into non-overlapping segments. For a frame group of size $G$, the $i$-th frame group contains frames $\mathbf{X}_{i} = \{\mathbf{x}_{iG}, \mathbf{x}_{iG+1}, \ldots, \mathbf{x}_{iG+G-1}\}$ for $i = 0, 1, 2, \ldots, \lceil T/G \rceil$.

Let $\mathbf{z}_{i,m} \in \mathbb{R}^{d_m \times T' \times W' \times H'}$ be the video latent embedding of frame group $i$ at encoder layer $m$, where $H', W'$ are the down-sampled spatial dimensions, $T'$ is the down-sampled temporal dimension, and $d_m$ is the latent dimension for layer $m$.

The causal caching mechanism ensures that the decoder ($\mathcal{D}$) for frame group $i$ receives context from the previous group. Specifically, let $\mathbf{z}_{i-1, m}^{c}$ denote the causal cache of size $N$ of decoded features from group $i-1$ for the decoder layer $m$. 
The decoder then reconstructs the frames and constructs the cache
\begin{equation}
\mathbf{z}_{i,m+1}, \mathbf{z}_{i,m}^c = \mathcal{D}_m(\text{Concat}(\mathbf{z}_{i-1, m}^{c}, \mathbf{z}_{i, m})).
\end{equation}

The causal structure creates a recurrent dependency where the pixel-wise loss $\mathcal{L}^{\text{pix}}_{i}$ for group $i$ depends on all previous groups through the concatenated context $\mathbf{z}_{i-1}^{c}$ at each decoder layer.

\subsection{Truncated Backpropagation and Locality} \label{subsec:trunc_back}
Truncated backpropagation leverages temporal locality to enable efficient training while preserving the essential temporal dependencies.
The following analysis focuses on causal caching within the decoder network.

Let $\mathbf{z}_{i} \in \mathbb{R}^{d}$ denote the unrolled latent, where the layer indices $m$ are omitted for notational convenience. 
Let $D(i,j)$ be a distance metric such that $D(i,j)=0$ if and only if $i$ and $j$ refer to latents belonging to the same frame group.
This index-based distance formalism allows us to reason about temporal proximity and the influence of one latent on another.

Let $J_{i,j}=\partial \mathbf{z}_i/ \partial \mathbf{z}_j \in \mathbb{R}^{d \times d}$ denote the Jacobian of latent $i$ with respect to latent $j$.
The scalar influence measure is then defined as
\begin{equation} \label{eq:influence}
    L_{i \leftarrow j}:= \lVert J_{i,j} \rVert,
\end{equation}
for a chosen matrix norm. 
This quantity captures the effect of latent $j$ on latent $i$ and is a vector-norm on a vector space.

Temporal locality is defined as the existence of constants $C, \alpha > 0$ such that the influence measure decays exponentially with distance
\begin{equation} \label{eq:locality}
    L_{i \leftarrow j} \leq C \cdot \exp(- \alpha D(i,j)).
\end{equation}

Intuitively, this means that a latent only meaningfully affects nearby latents in time. Using the chain rule, the gradient of the overall loss $\mathcal{L}$ with respect to a latent $\mathbf{z}_i$ decomposes as 
\begin{equation}
    \frac{\partial \mathcal{L}}{\partial \mathbf{z}_j} = \sum_{i} \frac{\partial \mathcal{L}}{\partial \mathbf{z}_i} \frac{\partial \mathbf{z}_i}{\partial \mathbf{z}_j} = \sum_{i} \frac{\partial \mathcal{L}}{\partial \mathbf{z}_i} J_{i,j}.
\end{equation}

Taking the norm of both sides and applying the triangle inequality,
\begin{equation} \label{eq:final_grad}
    \left\Vert \frac{\partial \mathcal{L}}{\partial \mathbf{z}_j} \right\Vert \leq \sum_{i} \left\Vert \frac{\partial \mathcal{L}}{\partial \mathbf{z}_i} \right\Vert L_{i \leftarrow j},
\end{equation}
which shows that the loss gradient at $\mathbf{z}_i$ is dominated by contributions from latents in close temporal proximity assuming temporal locality holds. Our key insight is that the temporal locality enables effective truncated backpropagation in the 3D VAE decoder. When we truncate gradients to only flow through a limited number of previous frame groups, the exponential decay in the influence measure ensures that the approximation error is bounded.

Specifically, for truncated backpropagation at temporal distance $D_{\text{trunc}}$, the error in gradient computation is bounded by
\begin{equation} \label{eq:full_error}
    \left\Vert \frac{\partial \mathcal{L}}{\partial \mathbf{z}_j} - \frac{\partial \mathcal{L}_{\text{trunc}}}{\partial \mathbf{z}_j} \right\Vert \leq C \cdot \exp(-\alpha D_{\text{trunc}}) \sum_{i} \left\Vert \frac{\partial \mathcal{L}}{\partial \mathbf{z}_i} \right\Vert,
\end{equation}

where $\mathcal{L}_{\text{trunc}}$ denotes the loss computed with truncated backpropagation.

A truncation distance $D_{\text{trunc}} \geq \frac{1}{\alpha} \log(\frac{C}{\epsilon})$ can therefore be chosen to satisfy a desired error tolerance $\epsilon$.
In practice, the network still learns effectively with a small truncation distance as shown in Sections \ref{subsec:analysis} and \ref{sec:apps}.

The integration of causal caching with truncated backpropagation creates a hybrid approach: the network backbone can still attend to all video latent embeddings for global temporal understanding, while the 3D VAE decoder operates with limited temporal context, reducing computational complexity. 
This design preserves essential temporal dependencies while making large-scale video diffusion model training using pixel-wise losses computationally tractable.

\subsection{Analysis}
\label{subsec:analysis}

\begin{figure}
    \centering
    \includegraphics[width=0.9\linewidth]{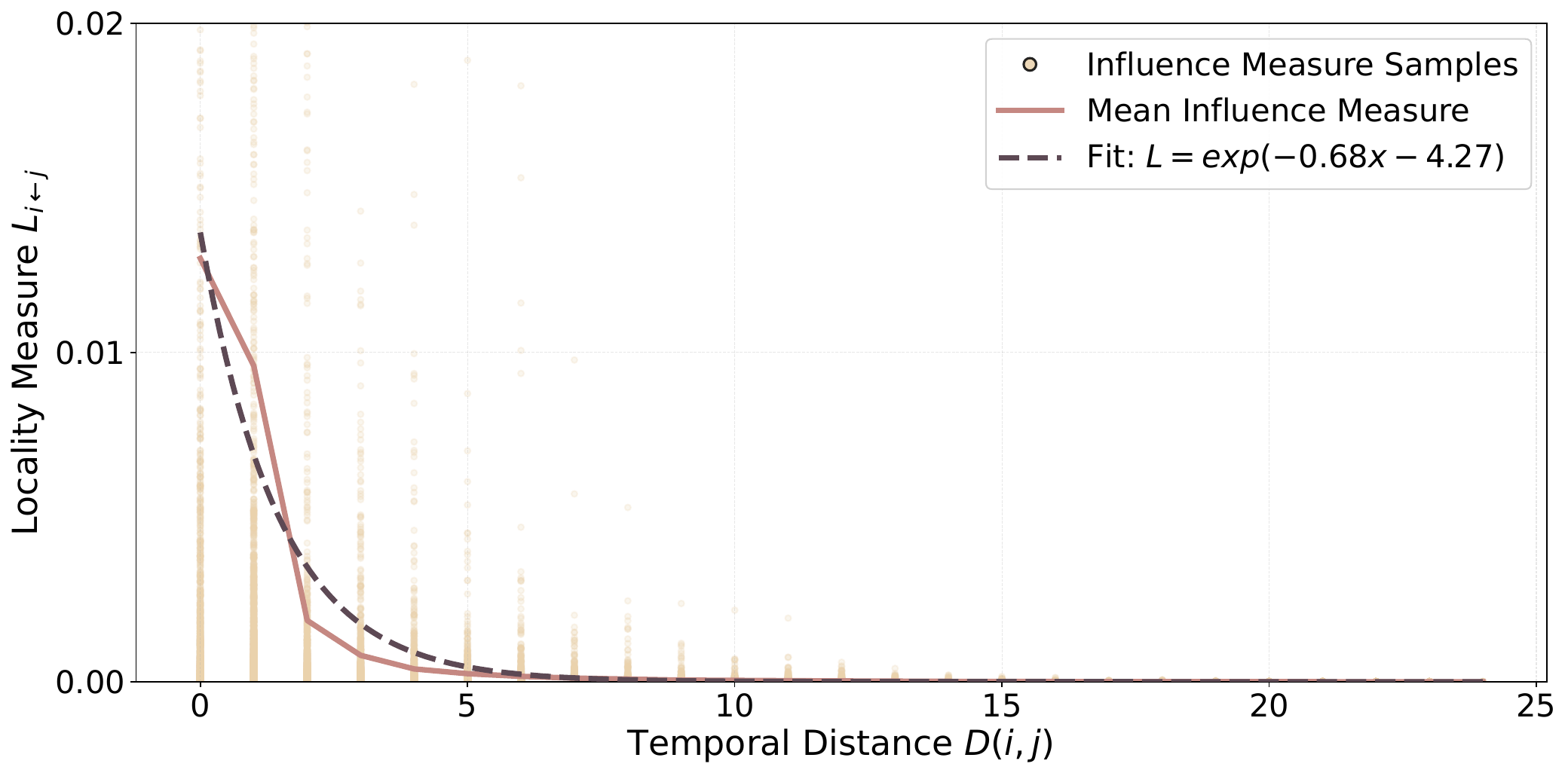}
    \caption{\textbf{Temporal Locality}. Influence measure samples \eqref{eq:influence} as a function of temporal distance between decoder inputs (i.e. latent embeddings) and outputs (i.e. pixels) alongside the mean and line of best fit. As temporal distance increases, the influence between embeddings decreases exponentially, resulting in minimal gradient contributions \eqref{eq:final_grad}.}
    \label{fig:influence_time}
\end{figure}
\begin{figure}
    \centering
    \includegraphics[width=0.9\linewidth]{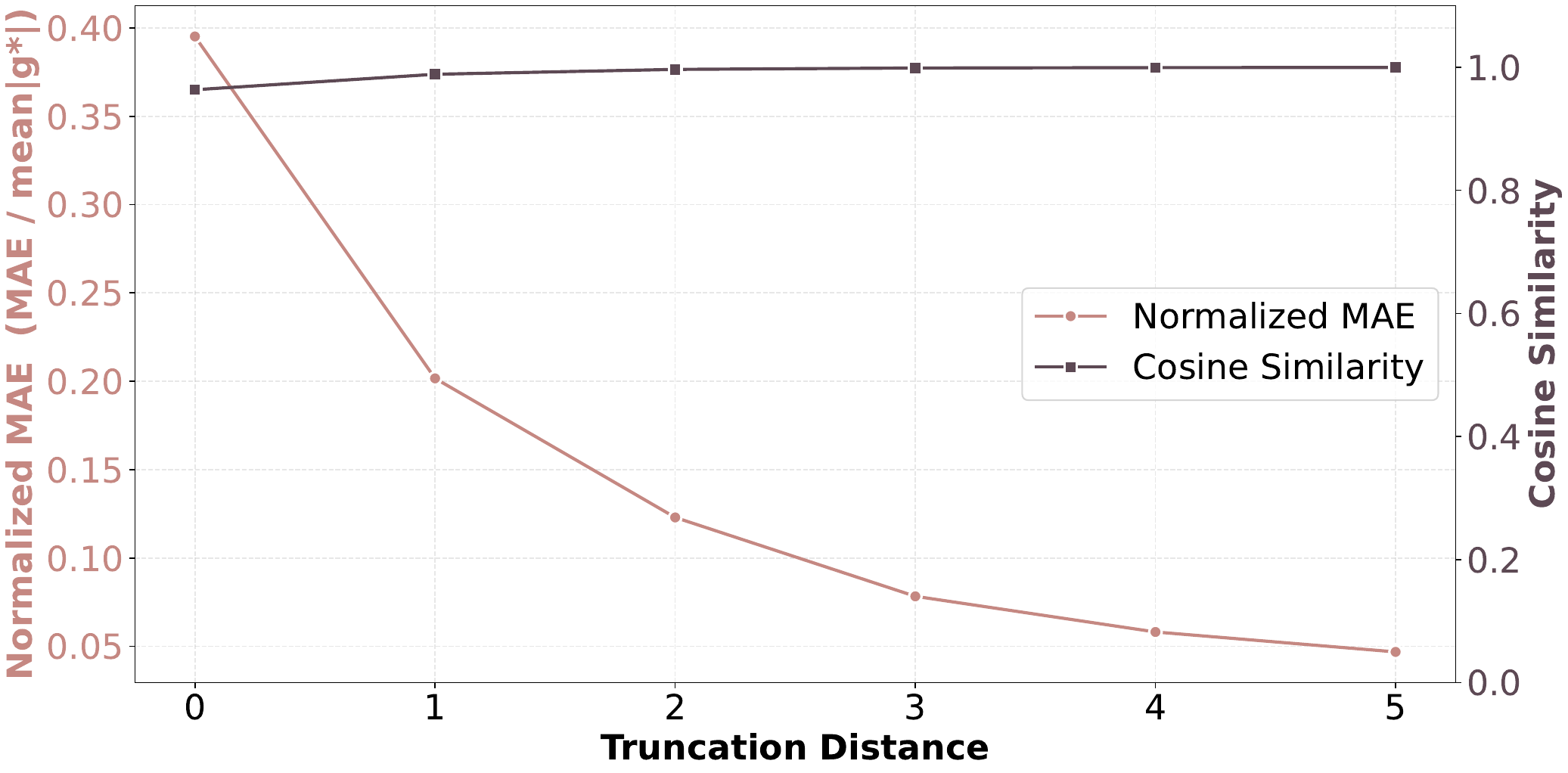}
    \caption{\textbf{Impact of Truncation Distance on Backbone Model Parameter Gradients}. Normalized MAE and cosine distance (computed by flattening all model parameters) are shown. Though error is significant at small truncation distances, the cosine similarity remains high across all distances, implying that the errors are primarily of magnitude, not direction.}
    \label{fig:transformer_grads}
\end{figure}

\paragraph{Temporal Locality.}
We analyze the proposed method by first confirming that temporal locality holds in the popular WAN 2.1 video decoder \cite{wan2025wan}.
The locality measure \eqref{eq:locality} is averaged across several videos, each with $97$ frames and down-sampled to a resolution of $64\times128$ to prevent prohibitive memory requirements.
Fig.~\ref{fig:influence_time} reports the mean of the influence measure \eqref{eq:influence} as a function of temporal distance, where a distance of $0$ indicates pixel $i$ is in the frame group of latent $j$.
Notably, the locality measure decays at an exponential rate, meaning the influence of pixels on frame groups significantly decreases as the temporal distance increases.
This property is demonstrated implicitly for other 3D VAEs by the results presented in Section \ref{sec:apps}.

\begin{figure}
    \centering
    \includegraphics[width=0.93\linewidth]{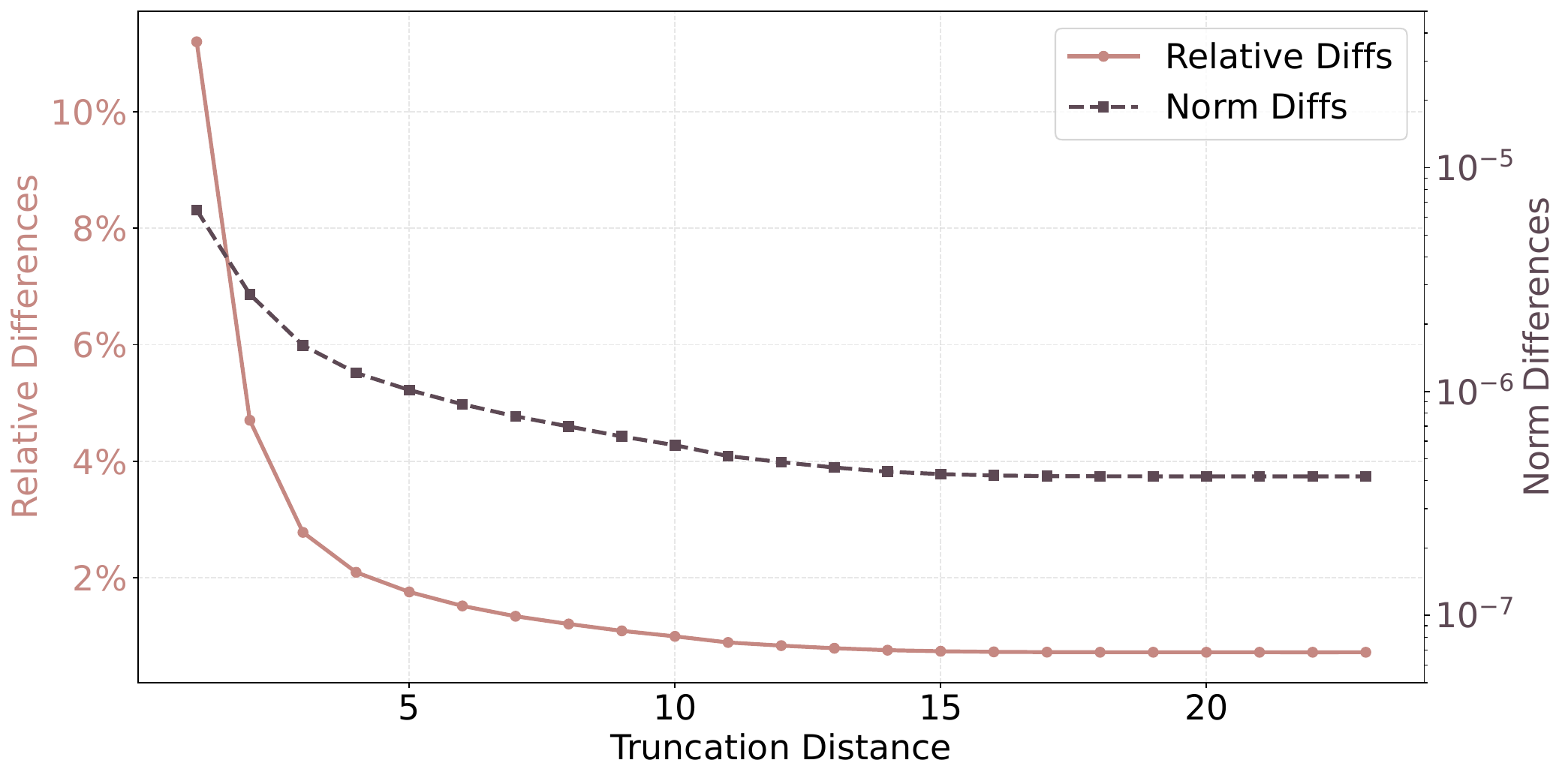}
    \caption{\textbf{Truncation Induced Gradient Error.} Mean gradient error \eqref{eq:full_error} between the truncated and full backpropagation algorithms as a function of truncation distance.}
    \label{fig:grad_accum}
\end{figure}

\paragraph{Decoder Input Gradient Error.}
We likewise present the gradient error \eqref{eq:full_error} between the full and truncated backpropagation algorithms as a function of truncation distance.
Gradients are computed by backpropagating pixel-wise losses to each decoder input latent considering varying truncation distances.
Reported results are the absolute and relative difference between the gradients for the truncated distance and the full backpropagation scheme.
Differences are measured using the Frobenius matrix norm and these, along with relative differences, are presented in Fig.~\ref{fig:grad_accum}.
From this plot we see that, even for low truncation distances, gradients approach those of full backpropagation, confirming that truncated backpropagation can be applied with minimal degradation in temporal consistency as the decoder network only considers small temporal neighborhoods.

\paragraph{Effect on Backbone Model Parameters.}
Next, we evaluate the effect of gradient truncation on the backbone model parameters during training by computing the average gradient of the parameters of the public Wan 2.1 1.3B transformer checkpoint over the entire training set of the DL3DV-benchmark dataset (see Section \ref{subsec:novel_view_synthesis}), around 100 videos. We perform this computation over a range of truncation distances and compare to the gradients of the full backwards pass, with results presented in Fig.~\ref{fig:transformer_grads}. 
Reported is the normalized mean absolute error (MAE) and cosine similarity, computed by flattening all model parameters into a single vector. 
The error is large for small truncation distances, indicating that the errors introduced by truncation are not averaged out over the dataset, and are propagated to model parameters. However, the high cosine similarity indicates that the error is primarily one of magnitude, not direction, and since gradient magnitudes are scaled by optimizers, the impact on training is negligible. 
This is confirmed by the results in Table \ref{tab:results_splat}, where increasing truncation distance only modestly improves performance.

\begin{figure}
    \centering
    \includegraphics[width=0.9\linewidth]{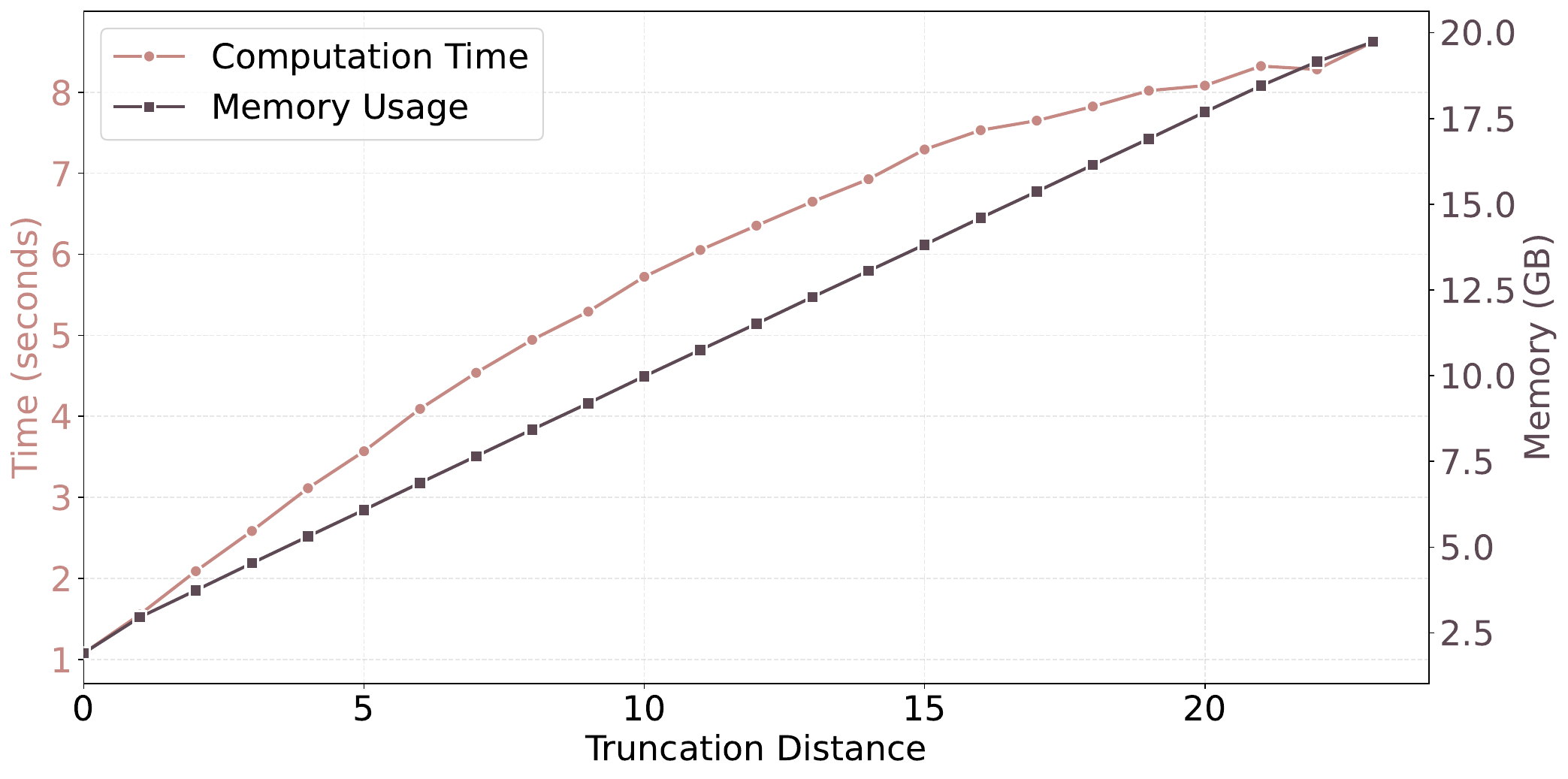}
    \caption{\textbf{Resource Utilization.} Computational time and memory requirements as a function of truncation distance.}
    \label{fig:time_and_memory}
\end{figure}

\begin{figure}
    \centering
    \includegraphics[width=\linewidth]{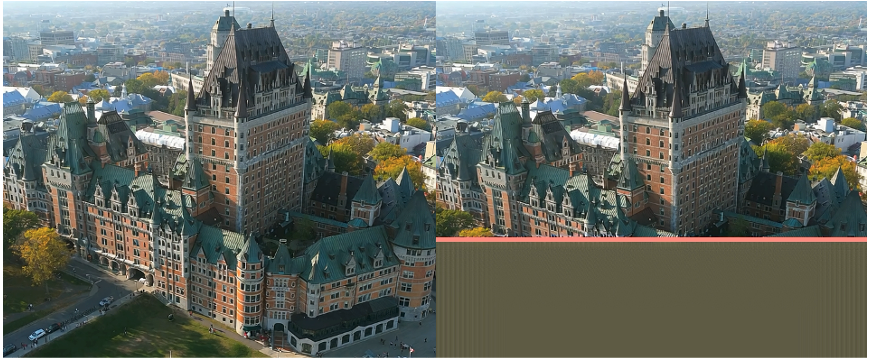}
    \caption{\textbf{Spatial Locality in 3D VAEs}. The video frame on the left is decoded from the original latents, while on the right a section of latents is zeroed. The red line indicates the boundary between original and zeroed latents. The upper portion of the frame is entirely unaffected by the corruption of the bottom.}
    \label{fig:spatial_locality}
\end{figure}

\paragraph{Runtime and Memory.}
Fig.~\ref{fig:time_and_memory} confirms that the proposed approach scales linearly with respect to truncation distance in terms of both computational time and memory. We reiterate that memory use is constant with respect to video length. To further save on memory, gradients are truncated spatially as well as temporally, such that gradients are computed over spatial chunks of the video separately. This spatial locality is illustrated in Fig.~\ref{fig:spatial_locality} and has been explored and leveraged by existing state-of-the-art video diffusion models \cite{wan2025wan, yang2024cogvideox}.

%% file: sections/04_Experiments.tex
\section{Applications} \label{sec:apps}

We validate the efficacy of \textit{ChopGrad} in four applications across multiple diffusion models: video super-resolution (\cref{subsec:super_resolution}),  novel view synthesis (\cref{subsec:novel_view_synthesis}), video inpainting (\cref{subsec:video_inpainting}), and controlled driving video generation (\cref{subsec:driving_video}).

\subsection{Video Super-Resolution}
\label{subsec:super_resolution}
\input{tables/super_res_comparisons}
We first show that adding \textit{ChopGrad} to a state-of-the-art video super-resolution method yields significant improvements in perceptual losses by finetuning DOVE \cite{chen2025dove} using \textit{ChopGrad}. DOVE finetunes CogVideoX \cite{yang2024cogvideox}, a DiT (Diffusion Transformer) model, for super-resolution. 
DOVE uses pixel-wise losses, including MSE and DISTS \cite{ding2020iqa}, but is forced to encode and decode each video frame separately during loss computation due to memory constraints, reducing inter-frame consistency and requiring the addition of a frame consistency loss to attempt to compensate for this. In contrast, for \textit{ChopGrad}, we start with the publicly available DOVE checkpoint and perform full finetuning on the HQ-VSR dataset \cite{chen2025dove} for 500 steps using video lengths of 24 frames, omitting interframe consistency losses. We use frame-wise DISTS loss with a weight of 0.1 and pixel-wise MSE with a weight of 1. All other settings are consistent with the original DOVE Stage-2 implementation, except that in DOVE 80\% of the batches are images, not videos, while we train on videos only. For the DOVE baseline, the publicly available DOVE checkpoint is used. As we found additional fine-tuning using the original DOVE method to result in equivalent performance, the results for the original model are presented.

\begin{figure}
    \centering
    \begin{minipage}{0.32\linewidth}\centering \scriptsize High-Res\end{minipage}
    \begin{minipage}{0.2\linewidth}\centering \scriptsize Low-Res\end{minipage}
    \begin{minipage}{0.2\linewidth}\centering \scriptsize DOVE\end{minipage}
    \begin{minipage}{0.2\linewidth}\centering \scriptsize ChopGrad\end{minipage}
    \includegraphics[width=\linewidth]{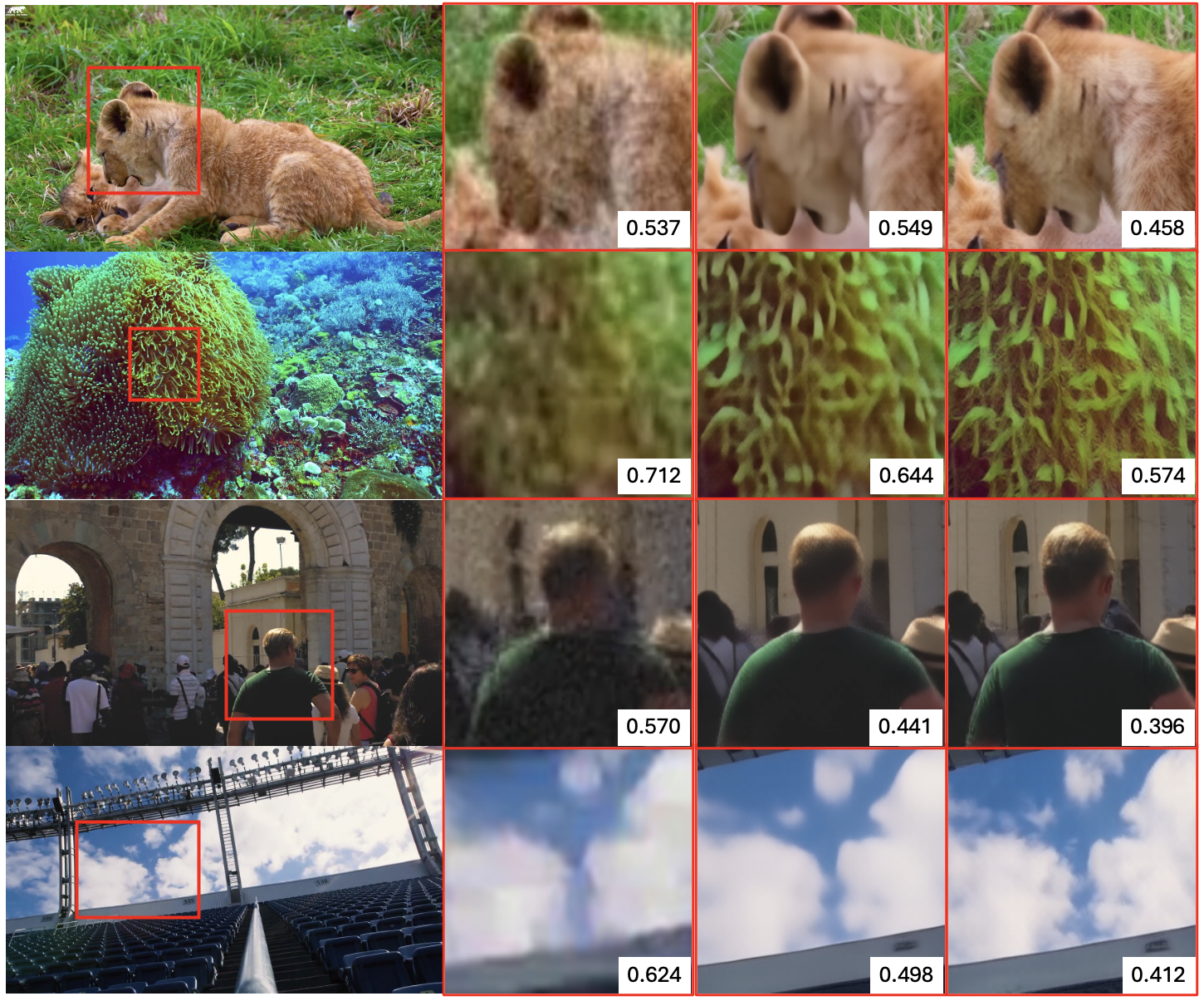}
    \caption{\textbf{Video Super-Resolution Comparison}. Shown from left to right: high-resolution, low-resolution input, DOVE \cite{chen2025dove}, and the proposed approach, \textit{ChopGrad}. \textit{ChopGrad} synthesizes fine textures better and reduces motion blur, especially in regions with high-frequency details like fur, hair, cloth, and clouds. LPIPS scores for each frame are shown in the bottom right-hand corner, where a lower score indicates better perceptual quality. The associated videos can be found in the Appendix.}
    \label{fig:super_res_examples}
    \vspace{-0.5cm}
\end{figure}

Quantitative results for video super-resolution are presented in Table \ref{tab:dove}. The addition of the proposed truncated backpropagation scheme improves performance across the majority of datasets and metrics, and the improvements are more pronounced for perceptual metrics (LPIPS and DISTS).
Selected frames from processed videos are shown in Fig.~\ref{fig:super_res_examples}, where \textit{ChopGrad} synthesizes fine-grained details such as fur, hair, and clouds better than the baseline approach.

\subsection{Artifact Removal in Novel View Synthesis}
\label{subsec:novel_view_synthesis}
Next, we use \textit{ChopGrad} for refining renders from imperfect neural rendering models \cite{kerbl20233d, mildenhall2021nerf}, which has recently become an established task \cite{wu2025difix3d, chen2024mvsplat360}.
Renders of 3D Gaussian Splatting novel view synthesis methods~\cite{kerbl20233d} often contain artifacts such as ``floaters'' that a set of recent diffusion models mitigate. 
Specifically, MVSplat-360 \cite{chen2024mvsplat360} and Difix3D+ \cite{wu2025difix3d} are designed for this task.
MVSplat-360 is trained to refine video sequences of 14 frames rendered from 3DGS models while Difix is trained to refine individual frames.
As a result, MVSplat-360 operates at a lower resolution ($448\times256$) with a small window of temporal consistency while Difix operates at a higher resolution ($960\times544$) but has no capacity to enforce temporal consistency.
While MVSplat-360 and Difix both leverage pixel-wise losses, they are unable to scale to long and high-resolution videos.

We generate a dataset using the DL3DV-Benchmark \cite{ling2024dl3dv}, a collection of 140 videos and camera trajectories. Gaussian splat models are generated using every $50$th frame of each video and rendered videos are constructed along entire camera trajectories. For \textit{ChopGrad}, we initialize the video diffusion model from a pre-trained Wan 2.1 14B \cite{wan2025wan} model and fine-tune the transformer backbone for 10 epochs.
Difix is fine-tuned for 10000 steps on the same data.
As MVSplat-360 is trained on the DL3DV dataset, no fine-tuning is applied.
We found that using the MVSplat-360 refinement model on our rendered videos led to poor performance.
Performance was significantly improved using the same number of sparse views for constructing the 3DGS model when using the views specified in the MVSplat-360 repository.
As such, we opt to use these improved selections for computing MVSplat-360 metrics.

\input{tables/3dgs_results}

\begin{figure*}
    \centering
    \begin{minipage}{0.98\linewidth}
        \begin{minipage}{0.19\linewidth}\centering \scriptsize Ground Truth\end{minipage}
        \begin{minipage}{0.19\linewidth}\centering \scriptsize 3DGS\end{minipage}
        \begin{minipage}{0.19\linewidth}\centering \scriptsize MVSplat-360\end{minipage}
        \begin{minipage}{0.19\linewidth}\centering \scriptsize Difix\end{minipage}
        \begin{minipage}{0.19\linewidth}\centering \scriptsize ChopGrad\end{minipage}
        \includegraphics[width=\linewidth]{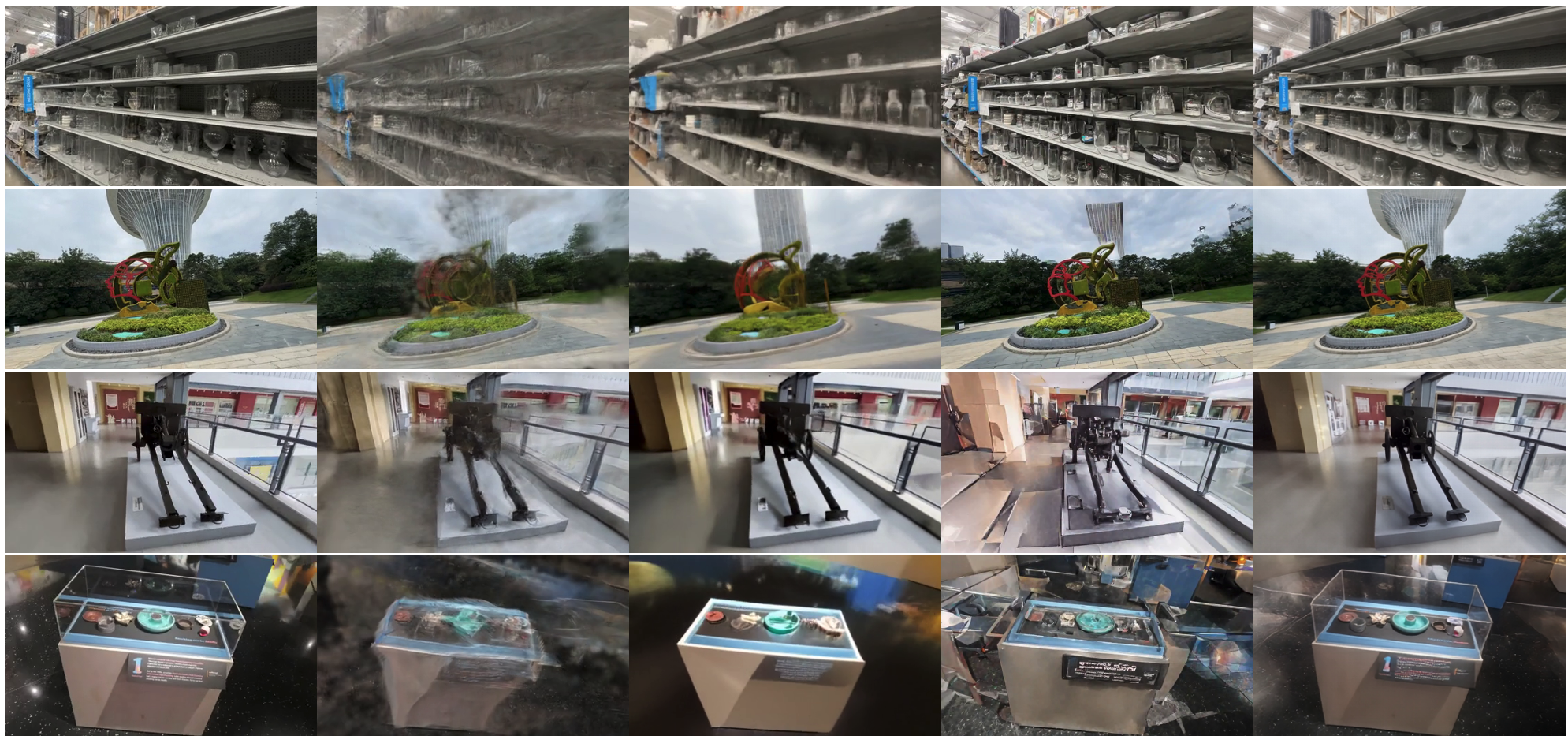}
    \end{minipage}
  
    \caption{\textbf{\textit{ChopGrad} vs Baselines for Neural Novel View Synthesis.} Ground truth video frames and 3D Gaussian Splat renders are shown on the left. Results for MVSplat-360 \cite{chen2024mvsplat360} and Difix \cite{wu2025difix3d} are presented alongside \textit{ChopGrad}.
    }
    \label{fig:3dgs_results}
\end{figure*}

\begin{figure}
    \centering
    \begin{minipage}{0.19\linewidth}\centering \scriptsize ChopGrad\textsuperscript{*}\end{minipage}
    \begin{minipage}{0.19\linewidth}\centering \scriptsize ChopGrad\textsuperscript{\textdagger}\end{minipage}
    \begin{minipage}{0.19\linewidth}\centering \scriptsize $D_{trunc}=0$\end{minipage}
    \begin{minipage}{0.19\linewidth}\centering \scriptsize $D_{trunc}=1$\end{minipage}
    \begin{minipage}{0.19\linewidth}\centering \scriptsize $D_{trunc}=2$\end{minipage}
    \includegraphics[width=\linewidth]{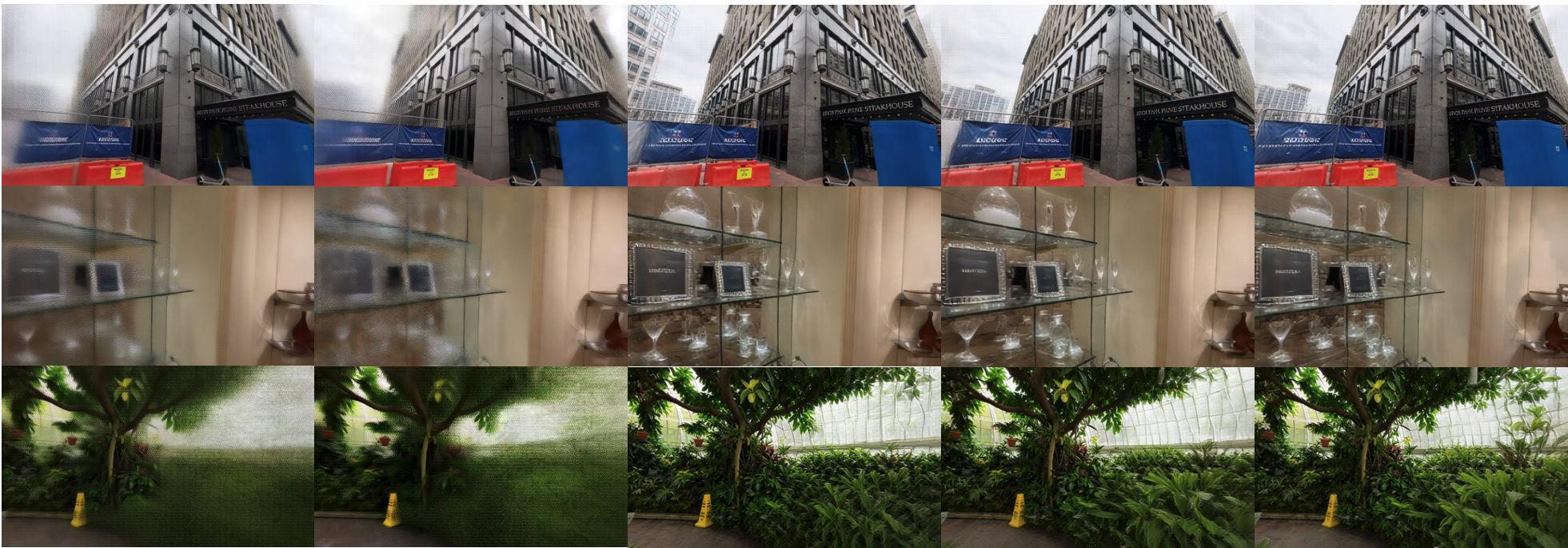}
    \caption{\textbf{Ablation Experiments for Neural Novel View Synthesis.} ChopGrad\textsuperscript{*} and ChopGrad\textsuperscript{\textdagger} are trained using only the MSE loss in the latent space. The $D_{trunc}$ cases show  \textit{ChopGrad} results at various truncation distances.}
    \label{fig:3dgs_ablation}
\end{figure}

Fig.~\ref{fig:3dgs_results} depicts \textit{ChopGrad} alongside the baseline methods for several scenes from the DL3DV-Benchmark test set and Table \ref{tab:results_splat} presents quantitative results. 
\textit{ChopGrad} out-performs the baselines across all metrics except temporal flickering where results are competitive with MVSplat-360.
A user study, available in the Appendix, also found that \emph{$95.6\%$ of users preferred the videos generated by ChopGrad} over those generated by MVSplat-360 or Difix.
Notably, while MVSplat-360 requires 60K training iterations \cite{chen2024mvsplat360}, \textit{ChopGrad} requires a small number of fine-tuning iterations when starting with the WAN2.1 14B pre-trained model. This demonstrates that \textit{ChopGrad} enables diffusion models to quickly generalize to unseen tasks by fine-tuning using pixel-space losses.

To demonstrate that the performance gains are a result of pixel-wise losses enabled by ChopGrad and not simply a more powerful backbone, we report ablation experiments in Table \ref{tab:results_splat} (bottom section) and a qualitative comparison in Fig.~\ref{fig:3dgs_ablation}, where \textit{ChopGrad} is trained using only MSE loss in the latent space and using various truncation distances.
While training only on the video latents is faster, the perceptual quality is worse and blurring is prevalent, especially in regions with fine details. As discussed in Section \ref{subsec:analysis}, truncation distance has a minor impact on result quality.
Videos of the DL3DV-Benchmark for \textit{ChopGrad} and baselines can be found in the Appendix.

\subsection{Video Inpainting}\label{subsec:video_inpainting}
We demonstrate that in video inpainting applications, \textit{ChopGrad} allows for reducing inference time by $50\times$ while remaining on-par in terms of quality. We evaluate \textit{ChopGrad} for video inpainting on three datasets: DL3DV-Benchmark \cite{ling2024dl3dv}, Waymo Open Dataset \cite{waymodataset}, and ROVI \cite{rovi}. For DL3DV-Benchmark and Waymo, we mask a fixed central region covering half the height and width of each frame and use an uninformative prompt. With ROVI, we use the included object masks and text descriptions. For \textit{ChopGrad} we finetune a Wan 2.1 14B model using latent MSE and pixel LPIPS losses for single-step inference using a truncation distance of 1. 
The baseline is VACE \cite{jiang2025vace} 14B, a control adapter for Wan 2.1 14B which is trained for a variety of tasks, including inpainting. VACE inference is performed using the default 50 steps from the VACE repository \cite{jiang2025vace}. For all datasets, we train both \textit{ChopGrad} and VACE the same number of steps. More training details are available in the Appendix.

\begin{figure}[t]
    \centering
    \begin{minipage}{0.95\linewidth}
        \begin{minipage}{0.32\linewidth}\centering \scriptsize VACE\end{minipage}
        \begin{minipage}{0.32\linewidth}\centering \scriptsize ChopGrad\end{minipage}
        \begin{minipage}{0.32\linewidth}\centering \scriptsize Ground Truth\end{minipage}
        \includegraphics[width=\linewidth]{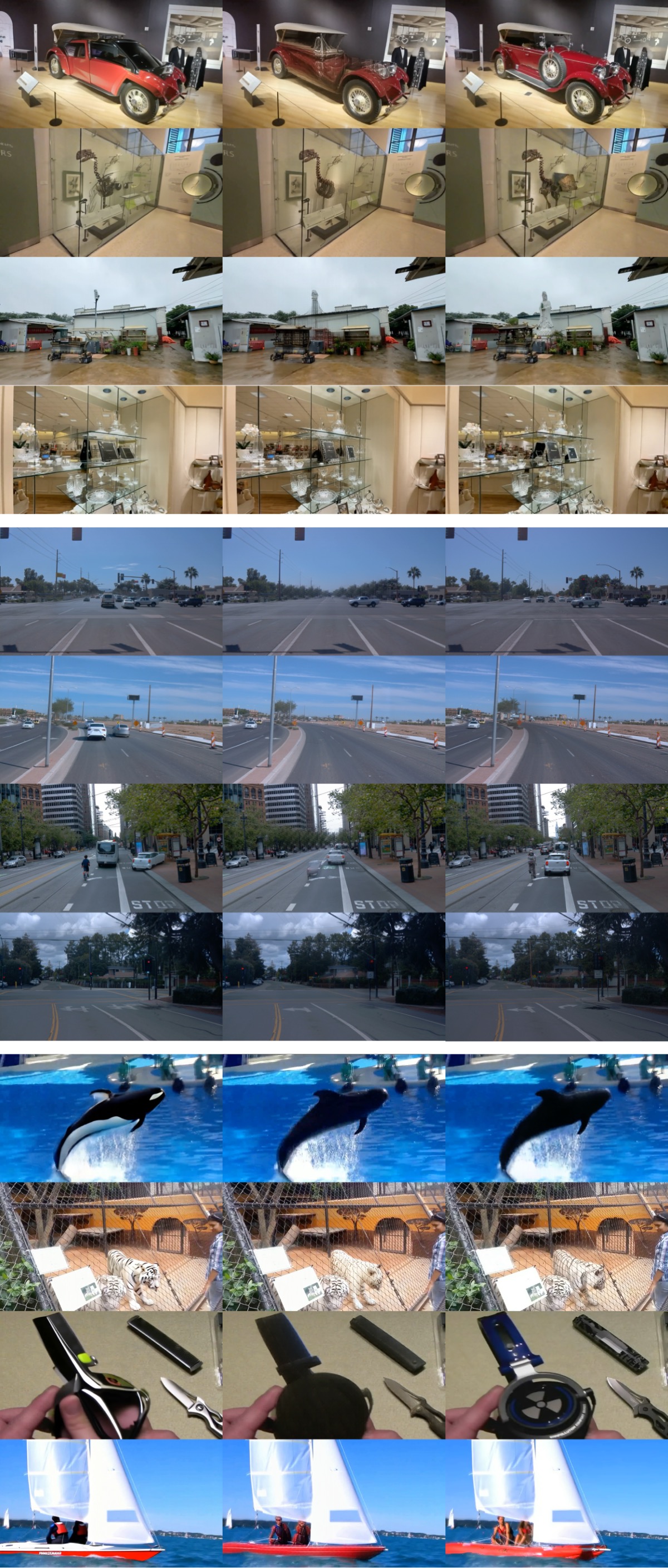}
    \end{minipage}
    \caption{\textbf{Video Inpainting}. We find that the recent VACE~\cite{jiang2025vace} tends to hallucinate (e.g., top section, top panel), while \textit{ChopGrad} stays closer to the input but can also produce implausible results. \textit{ChopGrad} results are output in a single step, a $50\times$ compute time improvement over VACE. Top:DL3DV, Middel: Waymo, Bottom: ROVI.}
    \label{fig:inpainting_results}
    \vspace{-0.5cm}
\end{figure}

Quantitative results are reported in Table \ref{tab:inpainting_quantitative}, qualitative results in Fig.~\ref{fig:inpainting_results}. \textit{ChopGrad} outperforms VACE on reconstruction-based metrics and maintains similar video quality metrics (VBench overall quality score within $1\%$ across all datasets) while reducing inference time compute budget by $50\times$. FVD (Fréchet Video Distance) is higher for \textit{ChopGrad} on ROVI but lower for the other two datasets, likely stemming from the overall more extreme masking in D3LDV and ROVI. Qualitatively, we observe that the \textit{ChopGrad} model adheres better to the scene and introduces fewer novel structures compared to VACE, occasionally at the cost of visual quality. In the more extreme masking regime of DL3DV and Waymo, VACE is penalized less for novel structures (relative to \textit{ChopGrad}), as the unmasked region is less informative about the region inside the mask, resulting in smaller relative improvements in reconstruction-based losses.

\subsection{Controlled Driving Video Generation}\label{subsec:driving_video}
\begin{figure}[t]
\centering
\begin{minipage}{0.24\linewidth}\centering \scriptsize Naive Insertion\end{minipage}
\begin{minipage}{0.24\linewidth}\centering \scriptsize Mirage\end{minipage}
\begin{minipage}{0.24\linewidth}\centering \scriptsize ChopGrad\end{minipage}
\begin{minipage}{0.24\linewidth}\centering \scriptsize Ground Truth\end{minipage}
 \includegraphics[width=\linewidth]{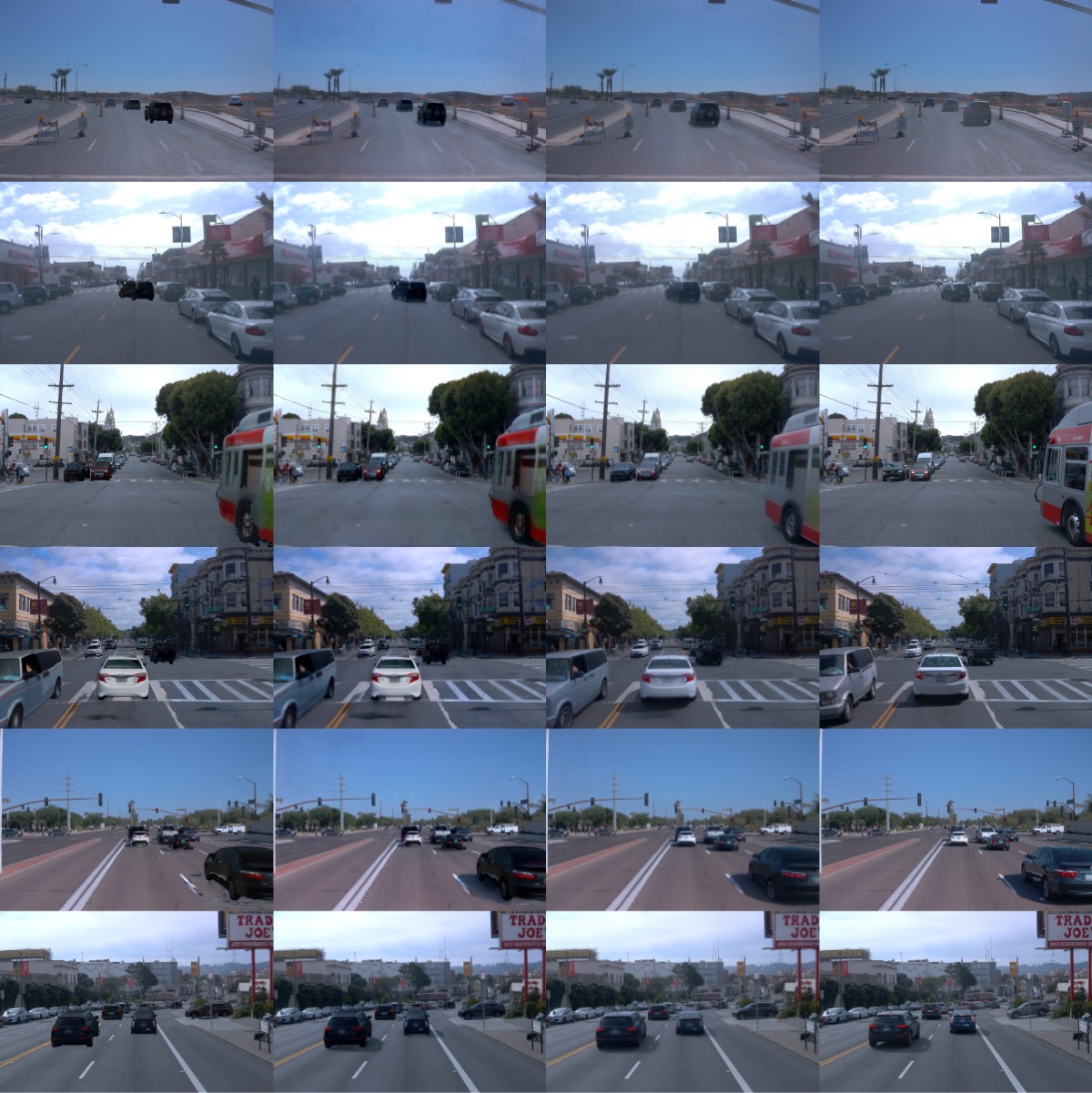}
    \caption{\textbf{Controlled Driving Video Generation}. Training with \textit{ChopGrad} improves lighting, removes more artifacts, and produces better shadows.}
    \label{fig:asset_variation_results}
    \vspace{-0.5cm}
\end{figure}

Visually realistic controlled driving video generation is essential for autonomous vehicle safety as it enables validation of vehicle behavior in rarely encountered scenarios. 3DGS~\cite{kerbl20233d} offers powerful scene reconstruction approaches, and recent neural driving simulators allow for manipulation of vehicles and reconstructed assets using scene graphs~\cite{ost2021neural, zhou2025hugsim, ljungbergh2024neuroncap} of reconstructed splats to enable this kind of simulation. However, large manipulation of vehicles and assets in these simulators~\cite{zhou2025hugsim, ljungbergh2024neuroncap} leads to myriad visual artifacts (see Naive Insertion columns of Fig.~\ref{fig:asset_variation_results} for examples). Post-processing videos rendered from such neural scenes with single-step diffusion is a promising approach for overcoming these issues, but existing methods such as \cite{ljungbergh2025r3d2,wang2025mirage} suffer from resolution / duration limitations.

\input{tables/inpainting}

Following \cite{ljungbergh2025r3d2, wang2025mirage} we create a dataset based on Waymo Open Dataset \cite{waymodataset} where 3DGS models are constructed, then assets are extracted and reinserted, producing the desired artifacts and input/output pairs to train and test with. Dataset construction details are presented in the Appendix.

We demonstrate \textit{ChopGrad} for controlled driving video generation on Mirage \cite{wang2025mirage} with our own Wan2.1-based implementation (details in the Appendix), 
as we were unable to acquire the original implementation even after contacting the authors. We train our implementation on 9-frame clips at a resolution of 480x832. After training Mirage we performed inference and evaluation at high resolution / duration (720x1280, 97 frames) as up-scaling training resolution outputs yielded poorer results. Subsequently, we finetuned Mirage's harmonization stage model using \textit{ChopGrad} for 1000 steps at 720x1280 resolution, 49 frame duration, and performed inference on 49 frame segments. Results are reported in Table \ref{tab:mirage_results} and Fig.~\ref{fig:asset_variation_results}. Quantitative metrics are improved across all tests, while inspection of the qualitative results shows that finetuning Mirage with ChopGrad improves lighting fixing, artifact removal, and shadow insertion. Notably, the parameters of the decoder itself are finetuned in Mirage, confirming that \textit{ChopGrad} can be used for decoder, as well as transformer, training.

\input{tables/asset_variation}

%% file: tables/super_res_comparisons.tex
\begin{table*}[t]
    \scriptsize
    \hspace{-2.mm}
    \centering
    \caption{\textbf{Quantitative Comparison for Video Super-Resolution}. The first, second, and third best results are highlighted with dark green, light green, and yellow, respectively. \textit{ChopGrad} outperforms all baselines in the majority of metrics and datasets, and achieves competitive performance otherwise.}
    \resizebox{0.95\linewidth}{!}{
        \setlength{\tabcolsep}{1.8mm}
        \begin{tabular}{l|l|c|c|c|c|c|c|c|c|c}
            \toprule[0.1em]
            \multirow{2}{*}{Dataset} & \multirow{2}{*}{Metrics} & RealESRGAN & ResShift & RealBasicVSR & Upscale-A-Video & MGLD-VSR & VEnhancer & STAR & DOVE & \textit{ChopGrad} \\
            & & \cite{wang2021real} & \cite{yue2023resshift} & \cite{chan2022investigating} & \cite{zhou2024upscale} & \cite{yang2024motion} & \cite{he2024venhancer} & \cite{xie2025star} & \cite{chen2025dove} & (Ours) \\
            \midrule
            \midrule
            \multirow{5}{*}{UDM10}
                 & PSNR ($\uparrow$) & 24.04 & 23.65 & 24.13 & 21.72 & \cellcolor{tabthird}24.23 & 21.32 & 23.47 & \cellcolor{tabsecond}26.48 & \cellcolor{tabfirst}26.70 \\
                 & SSIM ($\uparrow$) & \cellcolor{tabthird}0.7107 & 0.6016 & 0.6801 & 0.5913 & 0.6957 & 0.6811 & 0.6804 & \cellcolor{tabfirst}0.7827 & \cellcolor{tabsecond}0.7753 \\
                 & LPIPS ($\downarrow$) & 0.3877 & 0.5537 & 0.3908 & 0.4116 & \cellcolor{tabthird}0.3272 & 0.4344 & 0.4242 & \cellcolor{tabsecond}0.2696 & \cellcolor{tabfirst}0.2346 \\
                 & DISTS ($\downarrow$) & 0.2184 & 0.2898 & 0.2067 & 0.2230 & \cellcolor{tabthird}0.1677 & 0.2310 & 0.2156 & \cellcolor{tabsecond}0.1492 & \cellcolor{tabfirst}0.1143 \\
            \midrule
            \multirow{5}{*}{SPMCS}
                 & PSNR ($\uparrow$) & 21.22 & 21.68 & 22.17 & 18.81 & \cellcolor{tabthird}22.39 & 18.58 & 21.24 & \cellcolor{tabsecond}23.11 & \cellcolor{tabfirst}23.67 \\
                 & SSIM ($\uparrow$) & 0.5613 & 0.5153 & 0.5638 & 0.4113 & \cellcolor{tabthird}0.5896 & 0.4850 & 0.5441 & \cellcolor{tabsecond}0.6210 & \cellcolor{tabfirst}0.6274 \\
                 & LPIPS ($\downarrow$) & 0.3721 & 0.4467 & 0.3662 & 0.4468 & \cellcolor{tabthird}0.3263 & 0.5358 & 0.5257 & \cellcolor{tabsecond}0.2888 & \cellcolor{tabfirst}0.2647 \\
                 & DISTS ($\downarrow$) & 0.2220 & 0.2697 & 0.2164 & 0.2452 & \cellcolor{tabthird}0.1960 & 0.2669 & 0.2872 & \cellcolor{tabsecond}0.1713 & \cellcolor{tabfirst}0.1448 \\
            \midrule
            \multirow{5}{*}{YouHQ40}
                 & PSNR ($\uparrow$) & 22.82 & \cellcolor{tabthird}23.32 & 22.39 & 19.62 & 23.17 & 19.78 & 22.64 & \cellcolor{tabsecond}24.30 & \cellcolor{tabfirst}24.58 \\
                 & SSIM ($\uparrow$) & \cellcolor{tabthird}0.6337 & 0.6273 & 0.5895 & 0.4824 & 0.6194 & 0.5911 & 0.6323 & \cellcolor{tabsecond}0.6740 & \cellcolor{tabfirst}0.6760 \\
                 & LPIPS ($\downarrow$) & \cellcolor{tabthird}0.3571 & 0.4211 & 0.4091 & 0.4268 & 0.3608 & 0.4742 & 0.4600 & \cellcolor{tabsecond}0.2997 & \cellcolor{tabfirst}0.2581 \\
                 & DISTS ($\downarrow$) & 0.1790 & 0.2159 & 0.1933 & 0.2012 & \cellcolor{tabthird}0.1685 & 0.2140 & 0.2287 & \cellcolor{tabsecond}0.1477 & \cellcolor{tabfirst}0.1079 \\
            \midrule
            \multirow{5}{*}{RealVSR}
                 & PSNR ($\uparrow$) & 20.85 & 20.81 & \cellcolor{tabthird}22.12 & 20.29 & 22.02 & 15.75 & 17.43 & \cellcolor{tabsecond}22.32 & \cellcolor{tabfirst}22.43 \\
                 & SSIM ($\uparrow$) & 0.7105 & 0.6277 & \cellcolor{tabthird}0.7163 & 0.5945 & 0.6774 & 0.4002 & 0.5215 & \cellcolor{tabfirst}0.7301 & \cellcolor{tabsecond}0.7193 \\
                 & LPIPS ($\downarrow$) & 0.2016 & 0.2312 & \cellcolor{tabsecond}0.1870 & 0.2671 & 0.2182 & 0.3784 & 0.2943 & \cellcolor{tabfirst}0.1851 & \cellcolor{tabthird}0.1934 \\
                 & DISTS ($\downarrow$) & 0.1279 & 0.1435 & \cellcolor{tabthird}0.0983 & 0.1425 & 0.1169 & 0.1688 & 0.1599 & \cellcolor{tabsecond}0.0978 & \cellcolor{tabfirst}0.0944 \\
            \midrule
            \multirow{5}{*}{MVSR4x}
                 & PSNR ($\uparrow$) & \cellcolor{tabthird}22.47 & 21.58 & 21.80 & 20.42 & \cellcolor{tabfirst}22.77 & 20.50 & 22.42 & 22.42 & \cellcolor{tabsecond}22.55 \\
                 & SSIM ($\uparrow$) & 0.7412 & 0.6473 & 0.7045 & 0.6117 & 0.7418 & 0.7117 & \cellcolor{tabsecond}0.7421 & \cellcolor{tabsecond}0.7523 & \cellcolor{tabfirst}0.7550 \\
                 & LPIPS ($\downarrow$) & 0.4534 & 0.5945 & 0.4235 & 0.4717 & \cellcolor{tabthird}0.3568 & 0.4471 & 0.4311 & \cellcolor{tabsecond}0.3476 & \cellcolor{tabfirst}0.3212 \\
                 & DISTS ($\downarrow$) & 0.3021 & 0.3351 & 0.2498 & 0.2673 & \cellcolor{tabsecond}0.2245 & 0.2800 & 0.2714 & \cellcolor{tabthird}0.2363 & \cellcolor{tabfirst}0.2071 \\
            \bottomrule[0.1em]
        \end{tabular}
    }
    \label{tab:dove}
\end{table*}

%% file: tables/3dgs_results.tex
\begin{table*}[t]
\caption{\textbf{Neural Novel View Synthesis Results}. \textbf{Top section:} \textit{ChopGrad} out-performs all baselines across all metrics except temporal flickering, where it achieves competitive performance with MVSplat-360 \cite{chen2024mvsplat360}. Interestingly, while increasing the truncation distance noticeably increases training time memory, the metric differences are minimal. \textbf{Bottom section:} Ablation Results for \textit{ChopGrad}. \textit{ChopGrad}\textsuperscript{*} uses the same 1-step diffusion network, but is only trained using latent mean-squared error. \textit{ChopGrad}\textsuperscript{\textdagger} likewise uses latent mean-squared error for training but is trained twice as long. As such, both ablations do not propagate gradients through the video decoder. The performance of \textit{ChopGrad} using various truncation distances is also presented.
}
\scriptsize
\centering
\resizebox{0.95\linewidth}{!}{
\begin{tabular}{l|ccccccccc}
\toprule
\multirow{2}{*}{Method} & \multirow{2}{*}{FID ($\downarrow$)} & \multirow{2}{*}{PSNR ($\uparrow$)} & \multirow{2}{*}{SSIM ($\uparrow$)} & \multirow{2}{*}{LPIPS ($\downarrow$)} & \multirow{2}{*}{Dists ($\downarrow$)} & VBench Overall & VBench Temporal & Inference Time & Train Time \\
& & & & & & Quality ($\uparrow$) & Flickering ($\uparrow$) & [s/frame] & [H] \\
\midrule
\midrule
Difix \cite{wu2025difix3d} 
& \cellcolor{tabsecond}16.637 
& \cellcolor{tabsecond}17.213 
& \cellcolor{tabsecond}0.561 
& \cellcolor{tabsecond}0.407 
& \cellcolor{tabsecond}0.122 
& \cellcolor{tabsecond}0.766 
& \cellcolor{tabthird}0.898 
& \cellcolor{tabfirst}0.37 
& \cellcolor{tabfirst}2.0 \\

MVSplat-360 \cite{chen2024mvsplat360} 
& \cellcolor{tabthird}38.203 
& \cellcolor{tabthird}15.502 
& \cellcolor{tabthird}0.492 
& \cellcolor{tabthird}0.532 
& \cellcolor{tabthird}0.231 
& \cellcolor{tabthird}0.743 
& \cellcolor{tabfirst}0.926 
& \cellcolor{tabthird}2.89 
& - \\

\textit{ChopGrad} 
& \cellcolor{tabfirst}11.209 
& \cellcolor{tabfirst}19.237 
& \cellcolor{tabfirst}0.610 
& \cellcolor{tabfirst}0.342 
& \cellcolor{tabfirst}0.113 
& \cellcolor{tabfirst}0.783 
& \cellcolor{tabsecond}0.921 
& \cellcolor{tabsecond}1.11 
& \cellcolor{tabsecond}4.0 \\
\midrule
\textit{ChopGrad}\textsuperscript{*} & 48.525 & \cellcolor{tabfirst}19.501 & 0.588 & 0.440 & 0.244 & \cellcolor{tabthird}0.753 & \cellcolor{tabfirst}0.933 & 1.11 & \cellcolor{tabfirst}2.3\\
\textit{ChopGrad}\textsuperscript{\textdagger}  & 48.173 & \cellcolor{tabsecond}19.401 & 0.586 & 0.439 & \cellcolor{tabthird}0.238 & 0.751 & \cellcolor{tabsecond}0.932 & 1.11 & 4.5 \\
\textit{ChopGrad} $D_{trunc}=0$ & \cellcolor{tabthird}11.775 & 19.231 & \cellcolor{tabthird}0.605 & \cellcolor{tabthird}0.345 & \cellcolor{tabsecond}0.115 & \cellcolor{tabsecond}0.782 & 0.920 & 1.11 & \cellcolor{tabsecond}3.5 \\
\textit{ChopGrad} $D_{trunc}=1$ & \cellcolor{tabfirst}11.209 & 19.237 & \cellcolor{tabfirst}0.610 & \cellcolor{tabfirst}0.342 & \cellcolor{tabfirst}0.113 & \cellcolor{tabfirst}0.783 & 0.921 & 1.11 & \cellcolor{tabthird}4.0 \\
\textit{ChopGrad} $D_{trunc}=2$ & \cellcolor{tabsecond}11.742 & \cellcolor{tabthird}19.308 & \cellcolor{tabsecond}0.609 & \cellcolor{tabsecond}0.343 & \cellcolor{tabsecond}0.115 & \cellcolor{tabsecond}0.782 & \cellcolor{tabthird}0.922 & 1.11 & 4.5\\
\bottomrule
\bottomrule
\end{tabular}
}
\label{tab:results_splat}
\end{table*}

%% file: tables/inpainting.tex
\begin{table}[t]
\footnotesize
\centering
\caption{\textbf{Video Inpainting Evaluation}.
\textit{ChopGrad} results are output in a single step, a $50\times$ compute time improvement over VACE. VBench components are provided in the Appendix. Dark green is best, light green is second best.}
\label{tab:inpainting_quantitative}

\resizebox{\linewidth}{!}{
\begin{tabular}{llccccccc}
\toprule
\textbf{Dataset} & \textbf{Method} 
& \textbf{FID} $\downarrow$
& \textbf{FVD} $\downarrow$
& \textbf{PSNR} $\uparrow$
& \textbf{SSIM} $\uparrow$
& \textbf{LPIPS} $\downarrow$
& \textbf{DISTS} $\downarrow$
& \makecell{\textbf{VBench}\\\textbf{Overall}} $\uparrow$ \\
\midrule

\multirow{2}{*}{DL3DV}
& VACE     
& \cellcolor{tabsecond}45.060 
& \cellcolor{tabfirst}574.441 
& \cellcolor{tabsecond}20.678 
& \cellcolor{tabsecond}0.757 
& \cellcolor{tabsecond}0.236 
& \cellcolor{tabsecond}0.083 
& \cellcolor{tabfirst}0.792 \\
& ChopGrad 
& \cellcolor{tabfirst}40.948 
& \cellcolor{tabsecond}583.581 
& \cellcolor{tabfirst}21.699 
& \cellcolor{tabfirst}0.765 
& \cellcolor{tabfirst}0.221 
& \cellcolor{tabfirst}0.077 
& \cellcolor{tabfirst}0.792 \\
\midrule

\multirow{2}{*}{Waymo}
& VACE     
& \cellcolor{tabsecond}34.856 
& \cellcolor{tabfirst}440.651 
& \cellcolor{tabsecond}23.229 
& \cellcolor{tabsecond}0.804 
& \cellcolor{tabsecond}0.212 
& \cellcolor{tabsecond}0.079 
& \cellcolor{tabfirst}0.836 \\
& ChopGrad 
& \cellcolor{tabfirst}27.057 
& \cellcolor{tabsecond}470.545 
& \cellcolor{tabfirst}25.048 
& \cellcolor{tabfirst}0.823 
& \cellcolor{tabfirst}0.192 
& \cellcolor{tabfirst}0.071 
& \cellcolor{tabsecond}0.835 \\
\midrule

\multirow{2}{*}{ROVI}
& VACE     
& \cellcolor{tabsecond}30.491 
& \cellcolor{tabsecond}201.610 
& \cellcolor{tabsecond}22.618 
& \cellcolor{tabsecond}0.834 
& \cellcolor{tabsecond}0.223 
& \cellcolor{tabsecond}0.112 
& \cellcolor{tabfirst}0.752 \\
& ChopGrad 
& \cellcolor{tabfirst}27.547 
& \cellcolor{tabfirst}188.546 
& \cellcolor{tabfirst}25.200 
& \cellcolor{tabfirst}0.859 
& \cellcolor{tabfirst}0.199 
& \cellcolor{tabfirst}0.095 
& \cellcolor{tabsecond}0.747 \\
\bottomrule
\end{tabular}
}
\end{table}

%% file: tables/asset_variation.tex
\begin{table}[t]
\caption{\textbf{Controlled Driving Video Generation Results}. \textit{ChopGrad} was produced by initializing with Mirage followed by further finetuning Mirage's Harmonization stage for 1000 steps at high resolution (HR) / duration using \textit{ChopGrad}. Dark green is best, light green is second best.}
\centering
\small
\resizebox{\linewidth}{!}{
\begin{tabular}{lcccccc}
\hline
Method & PSNR $\uparrow$ & SSIM $\uparrow$ & LPIPS $\downarrow$ & DISTS $\downarrow$ & FID $\downarrow$ & FVD $\downarrow$ \\
\hline

Mirage (HR) 
& \cellcolor{tabsecond}27.30 
& \cellcolor{tabsecond}0.8912 
& \cellcolor{tabsecond}0.2067 
& \cellcolor{tabsecond}0.0740 
& \cellcolor{tabsecond}10.28 
& \cellcolor{tabsecond}204.66 \\

ChopGrad
& \cellcolor{tabfirst}29.49
& \cellcolor{tabfirst}0.9031
& \cellcolor{tabfirst}0.1719
& \cellcolor{tabfirst}0.0561
& \cellcolor{tabfirst}5.86
& \cellcolor{tabfirst}154.49 \\

\hline
\end{tabular}
}
\label{tab:mirage_results}
\end{table}

%% file: sections/05_Conclusion.tex
\section{Conclusion}
We introduce \textit{ChopGrad}, a truncated backpropagation approach that enables pixel-wise supervision at high resolutions and long durations in latent video diffusion models with causal caching. 
In architectures where the decoder is finetuned (e.g. \cite{wang2025mirage}) this capability is required, while in others it leads to significantly improved results (bottom of Table \ref{tab:results_splat}). Applications of such models trained with pixel-wise losses are numerous, including single-step model distillation \cite{dmd}, enhancement of neural rendered scenes \cite{wu2025difix3d, chen2024mvsplat360}, image translation \cite{pix2pixturbo}, video super-resolution \cite{chen2025dove}, and controlled driving video generation \cite{ljungbergh2025r3d2, wang2025mirage}.

By analyzing latent temporal locality, we demonstrate that long-range gradient dependencies in causal video autoencoders decay exponentially, allowing gradients to be truncated without compromising performance.
This insight enables efficient fine-tuning of high-resolution, long-duration video diffusion models using perceptual losses that were previously intractable due to recursive activation accumulation.

%% file: sections/06_Appendix_A.tex
\section{Implementation Details} \label{subsec:implem}
\textit{ChopGrad} is formalized in Algorithm \ref{alg:truncated_bp} and illustrated in Fig.2 of the primary document.
The latent cache is first either initialized as empty or detached from the previous decoder pass (lines \ref{alg:line:cache_detach} - \ref{alg:line:cache_compute}).
Critically, the cache is detached prior to running the forward pass, so the gradients do not propagate backwards through the full video.
The pixel-wise loss is computed using the decoded frames (line \ref{alg:line:loss}) and the gradients backpropagated to the latents and cache (lines \ref{alg:line:grad_first}-\ref{alg:line:grad_cache}).
Truncated backpropagation is then run using the specified truncation distance (lines \ref{alg:line:trunc_loop}-\ref{alg:line:trunc_grad_cache}), and the compute graph for latent $\mathbf{z}_{i-D_{trunc}}$ is subsequently released.
Cache gradients are zeroed after each backpropagation. 
Notably, maintaining the compute graph in memory requires storing activations, resulting in memory use that scales linearly with $D_{trunc}$, as does compute time as shown in Fig.6.

\input{algorithms/chopgrad_alg}
\vspace{-15pt}

%% file: algorithms/chopgrad_alg.tex
\begin{algorithm}[htb]
\caption{\textit{ChopGrad}.}
\label{alg:truncated_bp}
\begin{algorithmic}[1]
\REQUIRE Video latents $\{\mathbf{z}_{i}\}_{i=1}^{\lceil T/G \rceil}$, $D_{\text{trunc}}$, $\mathcal{L}_{pix}$.
\ENSURE Gradients $\{\nabla_{\mathbf{z}_{i}} \mathcal{L}\}$ for all latent frame groups
    
\FOR{$i$ from $1$ to $\lceil T/G \rceil$}
    \STATE $\mathbf{z}_{i-1}^{c} \leftarrow \text{detach}(\mathbf{z}_{i-1}^{c})$ \label{alg:line:cache_detach}
    
    \STATE $\hat{\mathbf{X}}_{i}, \mathbf{z}_{i}^{c} =\mathcal{D}(\text{Concat}(\mathbf{z}_{i-1}^{c}, \mathbf{z}_{i}))$\label{alg:line:cache_compute}
    
    \STATE $\mathcal{L}_{i} = \frac{1}{T} \sum_{t} \mathcal{L}^{pix}_i(\hat{\mathbf{X}}_{i,t}, \mathbf{X}_{i, t})$ \label{alg:line:loss}
    
    \STATE $\nabla_{\mathbf{z}_{i}} \leftarrow \frac{\partial \mathcal{L}_{i}}{\partial \mathbf{z}_{i}}$\label{alg:line:grad_first}
    
    \STATE $\nabla_{\mathbf{z}_{i-1}^{c}} \leftarrow \frac{\partial \mathcal{L}_{i}}{\partial \mathbf{z}_{i-1}^{c}}$ \label{alg:line:grad_cache}
    
    \FOR{$k = 1$ to $\min(D_{\text{trunc}}, i)$} \label{alg:line:trunc_loop}
        \STATE $\nabla_{\mathbf{z}_{(i-k)}} += \frac{\partial  \mathbf{z}_{(i-k)}^{c}}{\partial \mathbf{z}_{(i-k)}} \nabla_{\mathbf{z}_{(i-k)}^{c}}$ \label{alg:line:trunc_grad}
        
        \STATE $\nabla_{\mathbf{z}_{(i-k-1)}^{c}} \leftarrow \frac{\partial \mathbf{z}_{(i-k)}^{c}}{\partial \mathbf{z}_{(i-k-1)}^{c}}\nabla_{\mathbf{z}_{(i-k)}^{c}}$ \label{alg:line:trunc_grad_cache}
    \ENDFOR
    \IF{$i \geq D_{\text{trunc}}$}
        \STATE Release compute graph for $\mathbf{z}_{(i-D_{\text{trunc}})}$
    \ENDIF
\ENDFOR
\end{algorithmic}
\end{algorithm}

%% file: sections/07_Appendix_B.tex
\section{Additional Evaluation and Baseline Details}
\label{app:B}
This section provides additional evaluation and baseline details. \cref{subsec:video_super_res_baseline_appendix} presents details for Video Super-Resolution, \cref{subsec:novel_view_baseline_appendix} for Artifact Removal in Novel View Synthesis, \cref{subsec:appendix_baseline_video_inpainting} for Video Inpainting, and \cref{subsec:controlled_driving_baseline_appendix} for Controlled Driving Video Generation.
\subsection{Video Super-Resolution}
\label{subsec:video_super_res_baseline_appendix}
Video super-resolution is evaluated across the following datasets: UDM10 \cite{yi2019multi}, SPMCS \cite{tao2017spmc}, YouHQ40 \cite{zhou2024upscaleavideo}, RealVSR \cite{yang2021real}, and MVSR4x \cite{wang2023benchmark}.
Metric evaluations are performed using the publicly available DOVE evaluation script and reference baseline metrics are taken from those reported by DOVE.
Notably, the evaluation metrics computed using this script using the DOVE checkpoint match those originally reported.

\begin{figure}[h]
    \centering
    \includegraphics[width=0.7\linewidth]{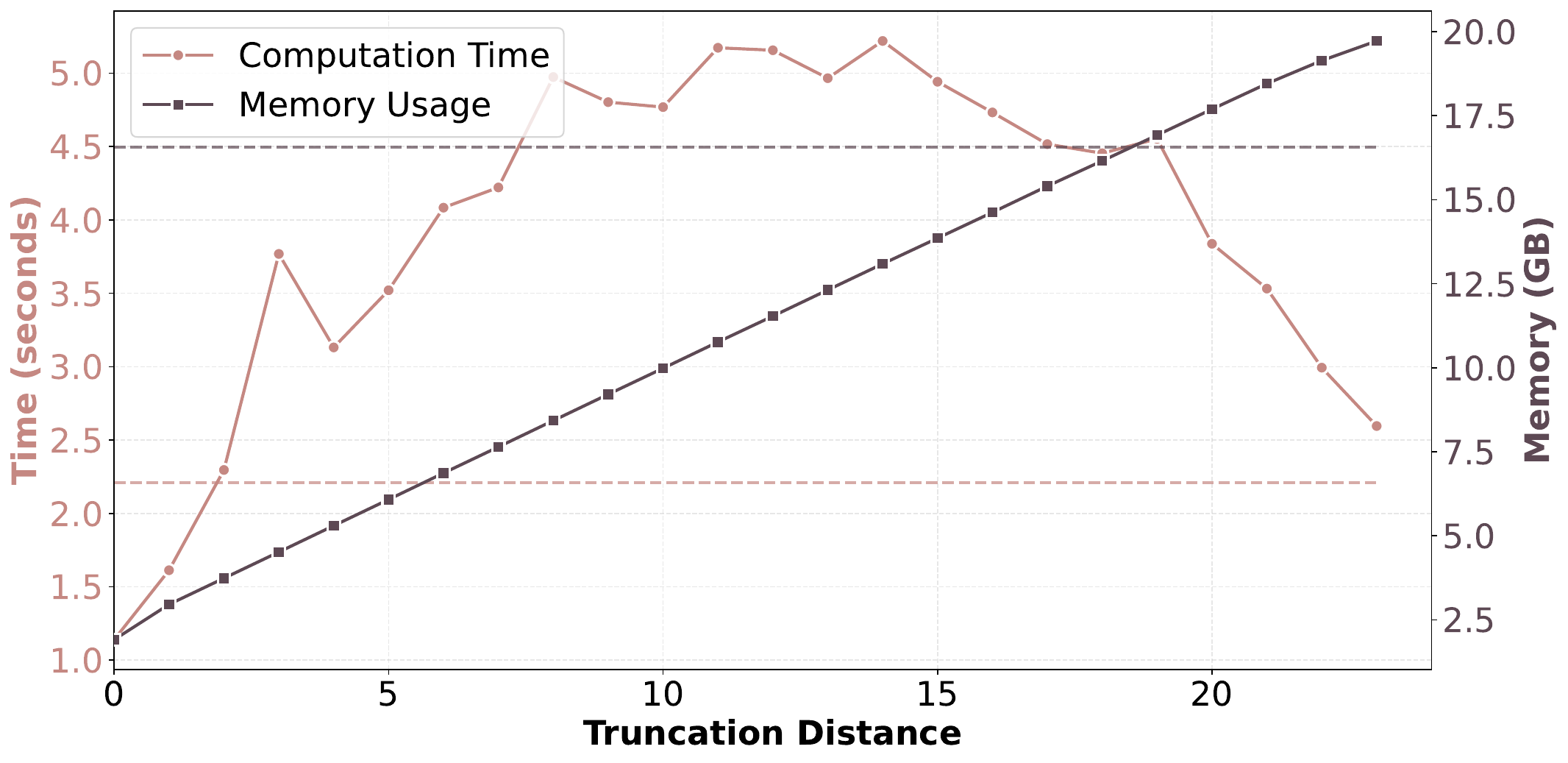}
        \caption{Computational time and memory requirements as a function of truncation distance for the modified \textit{ChopGrad} algorithm. The full video length is 24 frame groups. The dashed horizontal lines indicate time and memory requirements of the original backpropagation scheme. }
    \label{fig:modified_time_and_memory}
\end{figure}

\subsubsection{DOVE.}
The publicly available DOVE code is used for inference and evaluation. 
DOVE contains a 2-stage training scheme, where the first stage uses video sequences and the second stage uses a combination of video sequences and individual frames to train the network.

Additional Stage 2 fine-tuning was performed for $500$ iterations, but no performance improvement was observed, indicating that the public checkpoint was already converged and as such did not benefit from additional training. As such, the original DOVE checkpoint is used in the evaluations.

\subsection{Artifact Removal in Novel View Synthesis}
\label{subsec:novel_view_baseline_appendix}
We evaluate our neural-rendering enhancements on the DL3DV-Benchmark \cite{ling2024dl3dv}, which we split into 95 training scenes, 40 testing scenes, and 5 validation scenes. 
Each video in the dataset is approximately 300 frames long and a 3DGS model is trained using GSplat \cite{ye2025gsplat} using only every 50th frame for each scene.
Training of the 3DGS models follows the standard gsplat implementation, where the first 500 iterations do not modify the number of Gaussians, then the next 14.5k iterations are tailored for adaptive density control and the remaining iterations are for Gaussian parameter optimization.

The 3DGS model is trained using $960\times544$ resolution images for a total of 30k iterations, requiring approximately 5 minutes of training for each scene using an A100 GPU.

In addition, we conduct an anonymized study for determining user preference for novel view synthesis video generation.
Participants are presented with three side-by-side videos from ten scenes randomly selected from the test set. Each of the 3 videos is generated by one of the baselines, MVSplat \cite{chen2024mvsplat360}, Difix \cite{wu2025difix3d}, and the proposed method, ChopGrad ($D_{trunc}=1)$, and users are asked to mark the video which they preferred.
The order of video appearance for each scene was randomized.
A total of 34 users participated in the survey.
The percentage of users who preferred ChopGrad is computed for each scene and then averaged across all 10 scenes, yielding an overall preference rate of \emph{$95.6\%$} for the proposed method.

\subsubsection{Difix.}
Difix utilizes an image diffusion model backbone to process individual frames from input video sequences.
This diffusion network is initialized from the SD-Turbo \cite{sauer2024adversarial} checkpoint and is fine-tuned for neural render enhancement.

For a fair comparison, additional fine-tuning was performed on the DL3DV dataset using the training settings provided in the paper.
For each training iteration, video frames are randomly sampled from the dataset.
Difix is fine-tuned for $10$k iterations on 4 A100 GPUs, taking approximately $2$ hours.

\subsubsection{MVSplat360.}
The MVSplat-360 baseline approach takes as input a sparse set of views and uses a pre-trained feed-forward network to estimate a 3DGS model.
Next, a video sequence is generated along a camera trajectory and this video is refined using a fine-tuned video diffusion network.
The video diffusion network is initialized using the publicly available Stable Video Diffusion (SVD) checkpoint \cite{blattmann2023stable}.
The fine-tuned video diffusion checkpoint is provided by the MVSplat-360 authors and used in the experiments.

The SVD model groups together $14$ video frames during the diffusion process.
Self attention is used to enable frame groups to attend to one another, but this resulted in prohibitive memory requirements when evaluating videos with $294$ frames, even at low resolution.
As such, this feature was disabled and frame groups were not able to attend to each other.

As MVSplat-360 was trained on the DL3DV dataset, no additional fine-tuning was performed.

\subsection{Video Inpainting}
\label{subsec:appendix_baseline_video_inpainting}

We evaluate inpainting on 3 datasets: D3LDV-benchmark\cite{ling2024dl3dv}, Waymo\cite{waymodataset}, and ROVI\cite{rovi}. For D3LDV-Benchmark we use the same dataset and train/test split as was described in \cref{subsec:novel_view_baseline_appendix}. For Waymo, we use the dataset as described in \cref{subsec:controlled_driving_baseline_appendix}. For both of these cases we mask the ground truth videos as described in Section 4.2, with a fixed rectangular mask having half the height and half the width of the video. We also use a fixed prompt for both. We also include a new experiment where vehicle bounding boxes are masked in Waymo. Specifically, we randomly select 50\% of the labeled vehicles in the scene, and mask out the vehicle throughout the entire video. We refer to this setup as ``Waymo-Bbox.'' For ROVI we use the train/test split from the original dataset, as well as prompts provided by this dataset.

\subsubsection{VACE.}
For all experiments, we train the VACE baseline for the same number of steps as  \textit{ChopGrad}, with the same video duration, and resolution. All other settings are also the same, unless otherwise indicated in this section. VACE is trained using the standard latent MSE velocity objective (as Wan 2.1 and thus VACE are flow models), with the denoising timestep sampled using a timestep shift of 5, the same shift being used during sampling. The model is sampled over 50 steps following default settings in the VACE implementation\footnote{https://github.com/ali-vilab/VACE}. Masks are provided to the model during training and inference.

\subsection{Controlled Driving Video Generation}
\label{subsec:controlled_driving_baseline_appendix}
We evaluate on the Waymo Open Dataset \cite{waymodataset}, reconstructing 300 sequences, 230 for training and 70 for evaluation, using SplatAD \cite{hess2025splatad}, a dynamic 3D Gaussian Splatting-based method. SplatAD \cite{hess2025splatad} decomposes each scene into static background and dynamic actors represented as Gaussian primitives, which allows us to remove selected actors and replace them with generated 3D vehicles at the original pose. Our vehicle generation pipeline consists of vehicle extraction followed by vehicle alignment. We extract object-centric image patches from the curated Waymo object vehicle set using camera detections and LIDAR track IDs to assemble multi-view crops. Using instance and segmentation masks, we remove background pixels, and the resulting patches are used as input to the TRELLIS image→3D pipeline \cite{trellis} to produce 3D Gaussian representations of the vehicles. Since TRELLIS produces reconstructions in an unanchored coordinate frame, the generated vehicles can be arbitrarily rotated. To ensure consistent forward-facing poses, we render each vehicle from angles of 0 °, 90°, 180°, 270° and estimate its yaw with Orient‑Anything \cite{orient_anything}. vehicles with inconsistent cyclic orientation patterns or orientation confidence below a threshold are discarded, and the original 3DGS scene models are used.

\subsubsection{Mirage.}
\label{subsubsec:mirage_details}
Mirage code is not publicly available, and we were not able to gain access by emailing the authors, so we re-implement it based on the description of the method provided in the paper. We choose to use Wan 2.1 14B instead of CogVideoX as this is a more modern and capable diffusion model, but uses a similar VAE architecture. For fair comparison, for \textit{ChopGrad} we keep the same model architecture, losses, etc. and only modify the training duration and resolution, since use of \textit{ChopGrad} allows us to greatly increase these.

The main architectural modifications to the diffusion model proposed by Mirage are the addition of skip connections to the decoder, and the addition of several LoRAs. The skip connections are extracted from the encoder in ``2D'' mode, i.e., where each frame is encoded separately and no temporal compression is used. They are extracted from the output of the layer immediately preceding the first spatiotemporal downsample step, and are fused into the decoder immediately following the final spatio-temporal upsample step. To perform fusion, we concatenate the skip connections with decoder features along the channel dimension, then perform convolution with a $3\times3$ kernel to compute the fused features. We use LoRA rank and alpha values of 16 in the VAE, while for the transformer we set rank to 128 and alpha to 64.

We replicate Mirage's two training stages -- Reconstruction and Harmonization. We train each for $10k$ steps, with separate LoRAs for each stage. The Reconstruction phase trains the fusion blocks and a decoder reconstruction LoRA, while Harmonization trains a transformer LoRA and another decoder LoRA and keeps the fusion blocks frozen. For both stages learning rate is $2\times10^{-4}$, batch size is 8, clip length is 9 frames and clip resolution is $480\times832$. We updated the resolution slightly from the original paper to match Wan 2.1's training resolution. We trained the reconstruction phase with LPIPS and MSE losses, with the LPIPS component scaled by a factor of $0.1$.

The harmonization phase was trained with a fixed timestep (200). To be better aligned with Wan 2.1's velocity prediction training scheme, we trained the model to predict the difference between output and input. A combination of LPIPS and Gram losses was used, with Gram loss being scaled by $0.1$. The AdamW optimizer was used with a weight decay of 0.01, betas of 0.9 and 0.99. All trainings were performed on a node with 8 80GB A100 GPUs. Though the original paper trained on H200s with a smaller base model, we were able to fit our implementation on A100s by fully sharding the model and optimizer parameters with FSDP\footnote{https://docs.pytorch.org/docs/stable/fsdp.html}, and using transformer and decoder activation checkpointing.

%% file: sections/08_Appendix_C.tex
\section{Architecture, Training, and Inference Details}
\label{app:C}
This section provides additional details on architectures, training, and inference. \cref{subsec:appendix_super_res} presents details for Video Super-Resolution, \cref{subsec:supp_novel_view_synthesis} for Artifact Removal in Novel View Synthesis, \cref{subsec:supp_video_inpainting} for Video Inpainting, and \cref{subsec:appendix_controlled_driving} for Controlled Driving Video Generation.

\subsection{Video Super-Resolution}
\label{subsec:appendix_super_res}
\paragraph{Network Architecture.}
For video super-resolution, the DOVE \cite{chen2025dove} checkpoint is used to initialize \textit{ChopGrad}.
DOVE uses the CogVideoX \cite{yang2024cogvideox} autoencoder.
Similar to the WAN 2.1 \cite{wan2025wan} video autoencoder, the CogVideoX autoencoder compresses multiple video frames into frame groups.
This temporal compression differs from the WAN 2.1 autoencoder however, in that the number of decoded frames changes depending on how many frame groups are used.
When an odd number of frame groups are passed to the decoder, the first frame group is decoded into a single frame and the remaining frame groups are decoded into 4 frames.
When an even number of frame groups are passed to the decoder, all frame groups are decoded into 4 frames.
This behavior necessitates that 2 frame groups must be decoded together for each decoding step.
Although this behavior could be addressed through minor modifications to the implementation, no changes were made to preserve compatibility with the original network.

\paragraph{Training.}
The default DOVE configuration performs training at a resolution of $640\times320$.
Given this low resolution, no spatial chunking was used. 
Fine-tuning is conducted using videos with lengths of $24$ frames.
The second stage training implementation from DOVE is adopted with original hyperparameters and initialization is performed from the provided checkpoint.
Fine-tuning is performed for $500$ iterations on 4 A100 GPUs, requiring approximately $8$ hours. 
In contrast to the original DOVE Stage~2 procedure, only videos are used, no images are trained on.

\subsection{Artifact Removal in Novel View Synthesis}
\label{subsec:supp_novel_view_synthesis}
\paragraph{Network Architecture.} 
\textit{ChopGrad} and ablations are initialized using the pretrained Wan 2.1 14B \cite{wan2025wan} diffusion transformer model.
This model is then fine-tuned using the latent embeddings of the 3DGS renders as inputs and the ground-truth images as targets.
Notably, WAN is trained to output velocity $v = \hat{\boldsymbol{z}} - \boldsymbol{z}$, where $\hat{\boldsymbol{z}}$ is the latent embedding of the rendered video and $\boldsymbol{z}$ is the latent embedding of the ground truth video.
In order to better align with the original training objective of Wan 2.1 \cite{wan2025wan}, we leverage the same training scheme for fine-tuning.
A fixed text caption is used to condition the refinement and the diffusion timestep is fixed to 200.
No modifications to the video encoder have been made and the pre-trained WAN network has not been pre-distilled for single-step inference.

The WAN 2.1 video autoencoder has a temporal compression factor of $4$, meaning there are $4$ video frames per frame group latent.
Notably, the network pads the first video frame with $3$ empty frames, meaning the total length of the video processed by WAN 2.1 is $4N + 1$, where $N$ is the number of frame groups.
This temporal compression factor corresponds to $G=4$ from Section 3.4. \textit{ChopGrad} decodes latents to pixels using a spatial chunk size of $H/2 \times W/2$, resulting in 4 chunks.
This preserves the aspect ratio of the video and enables parallel processing for each chunk.

\paragraph{Training.}
\textit{ChopGrad} is trained using both a latent MSE loss and a pixel-space LPIPS \cite{zhang2018unreasonable} loss with VGG features.
An LPIPS weight of 100 was used while the latent MSE weight was set to 1 for all experiments, including ablations.

A scene is randomly chosen from the dataset and a random 81-frame sequence from this scene is used for each training iteration.
Videos have a resolution of $832\times480$.
Notably, larger resolution videos may be used for training (i.e. $1280\times720$), but lower resolution videos were used in experiments for fair baseline comparisons.

Training is conducted for approximately 3-4 hours on 8 A100 GPUs.
PyTorch's Fully Sharded Data Parallel architecture \cite{zhao2023pytorch} is leveraged to shard the model parameters and optimizer states of the WAN diffusion transformer.
To minimize memory, 8-way sequence parallelism is used.

The AdamW optimizer is used with a learning rate of $1e^{-5}$, a weight decay of $0.1$, betas of $0.9$ and $0.99$, and a batch size of $1$.
\textit{ChopGrad} and all ablations, excluding \textit{ChopGrad}\textsuperscript{\textdagger}, are trained for $880$ training iterations.
\textit{ChopGrad}\textsuperscript{\textdagger} is trained for $1760$ training iterations.
Training times are presented in Table 2.

\paragraph{Inference.}
Inference times in Table 1 are measured by processing the entire video and dividing by the number of total video frames and finally multiplied by the number of GPUs to account for sequence parallelism.

This ensures a fair comparison with the baseline methods which both utilize only a single GPU.
These inference times also include pre-processing as well as the full network pass (i.e., video encoding, decoding, and transformer forward pass).

Inference is performed on the first $297$ frames of the video as this corresponds to the temporal compression of the WAN 2.1 video autoencoder using a total of $N=75$ frame groups.
The evaluation is conducted on the first $294$ frames of the video to maintain compatibility and a fair comparison with the baseline methods as MVSplat-360 is only able to process multiples of 14 frames.

\subsection{Video Inpainting}
\label{subsec:supp_video_inpainting}

\paragraph{Network Architecture.} We use the same network setup as described in Section \ref{subsec:supp_novel_view_synthesis}. \textit{ChopGrad} is initialized using the pretrained Wan 2.1 14B \cite{wan2025wan} diffusion transformer model. This model is then fine-tuned using the latent embeddings of the masked videos as inputs and the ground-truth videos as targets. Notably, WAN is trained to output velocity $v = \hat{\boldsymbol{z}} - \boldsymbol{z}$, where $\hat{\boldsymbol{z}}$ is the latent embedding of the rendered video and $\boldsymbol{z}$ is the latent embedding of the ground truth video. In order to better align with the original training objective of Wan 2.1 \cite{wan2025wan}, we leverage the same training scheme for fine-tuning. A fixed text caption (except for the ROVI case) is used to condition the refinement and the diffusion timestep is fixed to 200. No modifications to the video encoder have been made and the pre-trained WAN network has not been pre-distilled for single-step inference.

The WAN 2.1 video autoencoder has a temporal compression factor of $4$, meaning there are $4$ video frames per frame group latent.
Notably, the network pads the first video frame with $3$ empty frames, meaning the total length of the video processed by WAN 2.1 is $4N + 1$, where $N$ is the number of frame groups.This temporal compression factor corresponds to $G=4$ from Section 3.4. \textit{ChopGrad} decodes latents to pixels using a spatial chunk size of $H/2 \times W/2$, resulting in 4 chunks. This preserves the aspect ratio of the video and enables parallel processing for each chunk.

\paragraph{Training.} Training is done using the same settings as described in Section \ref{subsec:supp_novel_view_synthesis}, except for differences in duration and number of training steps, which vary across the datasets. For DL3DV-benchmark we trained for 5 epochs at 49 frames, for Waymo (and Waymo-Bbox) 5 epochs at 49 frames, for ROVI $10k$ steps at 29 frames. The train times were 1.5, 3, and 18 hours, respectively. Train time for Waymo-Bbox was the same as for Waymo. The increased train steps for ROVI reflect the fact that it is a much larger dataset than the other two (5172 videos in the training set).

\paragraph{Inference.} We perform inference at the same duration as training, evaluating on the first N frames of each video, where N is the training/inference duration.

\subsection{Controlled Driving Video Generation}
\label{subsec:appendix_controlled_driving}

\paragraph{Network Architecture.} For \textit{ChopGrad}, we keep the same network as for Mirage (described in detail in Supplemental Section \ref{subsubsec:mirage_details}).

\paragraph{Training.} We initialize with the Mirage harmonization checkpoint which had been trained for $10k$ steps. The validation loss plateaued around $5k$ steps so we are confident the model had converged. We then train it for a subsequent $1k$ steps using \textit{ChopGrad} at a resolution of $720\times1280$ and a duration of 49 frames. We use 16 spatial chunks ($4h \times 4w$) and a truncation distance of 1. During this training batch size was set to 1 and the learning rate is maintained at $2\times10^{-4}$. The AdamW optimizer is used with a weight decay of 0.1, betas of 0.9 and 0.99. The training process took approximately 10 hours on 8 A100 GPUs.

%% file: sections/09_Appendix_D.tex
\section{Algorithmic Optimizations for Truncated Backpropagation}
The time complexity of \textit{ChopGrad} as described in Algorithm~\ref{alg:truncated_bp} scales linearly with respect to $D_{trunc}$. This complexity stems from the need to backpropagate over each frame group $D_{trunc}$ times. This is not the case with the standard backpropagation scheme, which only needs to make one backward pass, having accumulated gradients from all frame groups $i+1 ... N$ prior to computing the gradient for frame group $i$. This section examines in more detail the origin of this difference in time complexity and introduces a minor modification to Algorithm~\ref{alg:truncated_bp} that ensures, as the truncation distance approaches the full video length, the overall time complexity converges to that of the full backpropagation scheme.

\textit{ChopGrad} requires multiple backward passes over each frame group because in order to compute the gradient for frame group $i$, gradients for frame groups $i+1...i+D_{trunc}$ must be available. Furthermore, the compute graph for frame group $i$ cannot be released from memory until all $D_{trunc}$ future gradients have backpropagated through it. In order to release frame group $i$ as soon as possible, i.e. once the algorithm reaches frame group $i+D_{trunc}$, it is imperative to backpropagate all the way from $i+D_{trunc}$ to $i$ as soon $i+D_{trunc}$ is decoded and the loss computed. This necessitates performing a backward pass through all intermediate frame groups as well. Since backpropagation is performed through $D_{trunc}$ frame groups each time the compute graph from frame group $i$ is released, the compute must scale linearly with $D_{trunc}$. There is a time-memory tradeoff present -- graphs could be released less often, for example every $s$ steps instead of every single step, reducing the time complexity by a factor of $s$ but increasing memory consumption accordingly, since $D_{trunc} + s$ frames worth of activations need to be stored in memory.

In Algorithm~\ref{alg:truncated_bp}, backpropagation is performed all the way back through the previous $D_{trunc}$ steps as soon as a new frame group is decoded. If this is delayed so that the backpropagation only occurs when it is time to evict the compute graph for frame group $i-D_{trunc}$, a performance improvement can be gained in regimes where $D_{trunc}$ is close to the full video length, $T$. This is because the total number of backpropagation steps through individual frame groups (and as such the time complexity) becomes equal to $D_{trunc}*n_{evict}$, where $n_{evict}$ is the number of cache evictions, and $n_{evict} = T - D_{trunc}$. Empirical complexity results for this modification are shown in Figure \ref{fig:modified_time_and_memory}. Note that a low resolution video (128x64) is used to make computation of the vanilla backpropagation method and \textit{ChopGrad} with high truncation distances tractable.

In Figure \ref{fig:modified_time_and_memory} \textit{ChopGrad} has slightly worse time and memory performance that vanilla backpropagation at $D_{trunc}=T$ as some overhead is introduced by fragmenting the compute graph and maintaining detached versions of the cache.

It is not recommended to use \textit{ChopGrad} at truncation distances greater than 2, as the results from Section 4 demonstrate that causal VAEs exhibit strong spatial locality, and terms from faraway latents do not contribute significantly to the gradient. In regimes with small values of $D_{trunc}$, the modified and unmodified algorithms exhibit practically equivalent performance characteristics. This section is included for completeness, but should not have much impact on real-world uses of \textit{ChopGrad}.

%% file: sections/11_Appendix_F.tex
\section{Additional Quantitative Results}
Supplemental Table \ref{tab:inpainting_vbench_breakdown} presents individual VBench~\cite{huang2023vbench} component scores used to compile the VBench Overall Quality metric reported for the Video Inpainting application in main document Table 3. Higher is better for all scores. In addition to a $50\times$ reduction in compute time, \textit{ChopGrad} delivers modest improvements in motion smoothness and temporal flickering across all datasets, while VACE consistently has slightly higher imaging quality and subject consistency, with the remaining scores having mixed results across datasets. This pattern is consistent with our observation that VACE is more prone to hallucination -- conforming less closely to the input video allows it to generate slightly higher quality images (despite being included in VBench, aesthetic quality is an image-based metric) and more consistent subjects. The \textit{ChopGrad} trained single-step model hallucinates less, retains higher reconstruction metrics (as evidenced in main document Table 3), and its improved motion smoothness and temporal flickering can be attributed to it staying closer to the original video, as real videos often have smooth motion and less flicker than generated videos.

Supplemental Table~\ref{tab:waymo_bbox_quantitative} presents the same metrics as main document Table 3 for the new Waymo-Bbox task. Results are consistent with other tasks, with \textit{ChopGrad} training resulting in improved reconstruction metrics and similar video quality metrics, while achieving a $50\times$ inference time reduction.

\input{tables/supp_tables}

%% file: tables/supp_tables.tex
\begin{table*}
\footnotesize
\centering
\caption{\textbf{Breakdown of VBench Scores for Video Inpainting}. Full VBench component scores corresponding to the VBench Overall values reported in main document  Table 3. In addition to a $50\times$ reduction in compute time, \textit{ChopGrad} delivers modest improvements in motion smoothness and temporal flickering across all datasets, while VACE consistently has slightly higher imaging quality and subject consistency, with the remaining scores having mixed results across datasets.}
\label{tab:inpainting_vbench_breakdown}

\begin{adjustbox}{width=\textwidth}
\begin{tabular}{llcccccccc}
\toprule
Dataset & Method
& \makecell{Aesthetic\\Quality}
& \makecell{Background\\Consistency}
& \makecell{Dynamic\\Degree}
& \makecell{Imaging\\Quality}
& \makecell{Motion\\Smoothness}
& \makecell{Subject\\Consistency}
& \makecell{Temporal\\Flickering}
& \makecell{VBench Overall\\Quality} \\
\midrule

\multirow{2}{*}{DL3DV}
& VACE
& 0.531 & 0.917 & 0.950 & 0.731 & 0.957 & 0.912 & 0.912 & 0.792 \\
& ChopGrad
& 0.519 & 0.912 & 0.950 & 0.729 & 0.961 & 0.903 & 0.919 & 0.792 \\

\midrule

\multirow{2}{*}{Waymo}
& VACE
& 0.512 & 0.956 & 0.882 & 0.716 & 0.987 & 0.953 & 0.969 & 0.836 \\
& ChopGrad
& 0.517 & 0.960 & 0.894 & 0.695 & 0.988 & 0.945 & 0.972 & 0.835 \\

\midrule

\multirow{2}{*}{Waymo-Bbox}
& VACE
& 0.519 & 0.957 & 0.859 & 0.704 & 0.987 & 0.950 & 0.970 & 0.834 \\
& ChopGrad
& 0.525 & 0.962 & 0.847 & 0.696 & 0.988 & 0.955 & 0.972 & 0.835 \\

\midrule

\multirow{2}{*}{ROVI}
& VACE
& 0.472 & 0.926 & 0.816 & 0.538 & 0.959 & 0.885 & 0.941 & 0.752 \\
& ChopGrad
& 0.457 & 0.924 & 0.749 & 0.521 & 0.964 & 0.884 & 0.947 & 0.747 \\

\bottomrule
\end{tabular}
\end{adjustbox}
\end{table*}

\begin{table*}
\footnotesize
\centering
\caption{\textbf{Waymo-Bbox Video Inpainting Results}. Quantitative comparison between VACE and \textit{ChopGrad} on the Waymo-Bbox setting. Results are consistent with other tasks, with \textit{ChopGrad} training resulting in improved reconstruction metrics and similar video quality metrics, while achieving a $50\times$ inference time reduction.}
\label{tab:waymo_bbox_quantitative}

\begin{adjustbox}{width=0.6\linewidth}
\begin{tabular}{lccccccc}
\toprule
\textbf{Method} 
& FID
& FVD
& PSNR
& SSIM
& LPIPS
& DISTS
& VBench Overall \\
\midrule
VACE     
& \textbf{12.084} 
& 253.710 
& 27.661 
& 0.873 
& 0.154 
& 0.061 
& 0.834 \\
ChopGrad 
& 12.358 
& \textbf{252.599} 
& \textbf{28.838} 
& \textbf{0.875} 
& \textbf{0.146} 
& \textbf{0.057} 
& \textbf{0.835} \\
\bottomrule
\end{tabular}
\end{adjustbox}
\end{table*}

%% file: sections/10_Appendix_E.tex
\section{Additional Qualitative Results}
Additional qualitative results for Video Super-Resolution, Artifact Removal in Novel View Synthesis, and Controlled Driving Video Generation experiments are presented in Figures \ref{fig:super_res_appendix}, \ref{fig:3dgs_appendix}, and \ref{fig:asset_variation_appendix} respectively.
Qualitative results for inpainting are presented in four separate figures based on dataset: DL3DV in Fig.~\ref{fig:inpainting_appendix_d3ldv}, Waymo in Fig.~\ref{fig:inpainting_appendix_waymo}, Waymo-Bbox in Fig.~\ref{fig:inpainting_appendix_waymo_bbox}, and ROVI in Fig.~\ref{fig:inpainting_appendix_rovi}.
These additional results report a number of the benefits of training at increased resolution / duration with \textit{ChopGrad}. In the case of Video Super-Resolution (Fig. \ref{fig:super_res_appendix}), \textit{ChopGrad} enabled back-propagating through the decoder for entire videos (rather than individual frames), allowing the transformer to properly account for temporal compression and resulting in improved visual quality, especially for fine details such as fur, cloth, and clouds. In the case of Artifact Removal in Novel View Synthesis (Fig. \ref{fig:3dgs_appendix}), \textit{ChopGrad} trained models have access to more views of the scene simultaneously as a result of extended video duration, leading to enhanced artifact removal capabilities. In Video Inpainting (Figs. \ref{fig:inpainting_appendix_d3ldv}, \ref{fig:inpainting_appendix_waymo}, \ref{fig:inpainting_appendix_waymo_bbox}, and \ref{fig:inpainting_appendix_rovi}), \textit{ChopGrad} is used to produce a significantly faster model (single-step vs 50 steps for VACE) while also reducing hallucinations. Finally, Fig. \ref{fig:asset_variation_appendix} shows that for Controlled Driving Video Generation, tuning with \textit{ChopGrad} (as compared to training at low resolution/duration and performing high resolution/duration inference) leads to a stronger model that is able to make larger changes to input images, improving lighting and shadows and removing more Gaussian Splat artifacts. Collectively, these results illustrate a variety of ways in which state-of-the-art methods may be improved further by using truncated backpropagation to enable pixel-wise perceptual losses.

%% file: main.bib
@inproceedings{aicher2020,
  title={Adaptively truncating backpropagation through time to control gradient bias},
  author={Aicher, Christopher and Foti, Nicholas J and Fox, Emily B},
  booktitle={Uncertainty in Artificial Intelligence},
  pages={799--808},
  year={2020},
  organization={PMLR}
}

@article{yu2024viewcrafter,
  title={View{C}rafter: {T}aming video diffusion models for high-fidelity novel view synthesis},
  author={Yu, Wangbo and Xing, Jinbo and Yuan, Li and Hu, Wenbo and Li, Xiaoyu and Huang, Zhipeng and Gao, Xiangjun and Wong, Tien-Tsin and Shan, Ying and Tian, Yonghong},
  journal={arXiv preprint arXiv:2409.02048},
  year={2024}
}

@inproceedings{wu2025improved,
  title={Improved video {VAE} for latent video diffusion model},
  author={Wu, Pingyu and Zhu, Kai and Liu, Yu and Zhao, Liming and Zhai, Wei and Cao, Yang and Zha, Zheng-Jun},
  booktitle={Proceedings of the Computer Vision and Pattern Recognition Conference},
  pages={18124--18133},
  year={2025}
}

@article{yang2024cogvideox,
  title={Cog{V}ideo{X}: {T}ext-to-video diffusion models with an expert transformer},
  author={Yang, Zhuoyi and Teng, Jiayan and Zheng, Wendi and Ding, Ming and Huang, Shiyu and Xu, Jiazheng and Yang, Yuanming and Hong, Wenyi and Zhang, Xiaohan and Feng, Guanyu and others},
  journal={arXiv preprint arXiv:2408.06072},
  year={2024}
}

@inproceedings{li2025wf,
  title={{WF-VAE}: {E}nhancing video {VAE} by wavelet-driven energy flow for latent video diffusion model},
  author={Li, Zongjian and Lin, Bin and Ye, Yang and Chen, Liuhan and Cheng, Xinhua and Yuan, Shenghai and Yuan, Li},
  booktitle={Proceedings of the Computer Vision and Pattern Recognition Conference},
  pages={17778--17788},
  year={2025}
}

@article{wan2025wan,
  title={{WAN}: {O}pen and advanced large-scale video generative models},
  author={Wan, Team and Wang, Ang and Ai, Baole and Wen, Bin and Mao, Chaojie and Xie, Chen-Wei and Chen, Di and Yu, Feiwu and Zhao, Haiming and Yang, Jianxiao and others},
  journal={arXiv preprint arXiv:2503.20314},
  year={2025}
}

@article{ho2022video,
  title={Video diffusion models},
  author={Ho, Jonathan and Salimans, Tim and Gritsenko, Alexey and Chan, William and Norouzi, Mohammad and Fleet, David J},
  journal={Advances in neural information processing systems},
  volume={35},
  pages={8633--8646},
  year={2022}
}

@article{ho2022imagen,
  title={Imagen video: {H}igh definition video generation with diffusion models},
  author={Ho, Jonathan and Chan, William and Saharia, Chitwan and Whang, Jay and Gao, Ruiqi and Gritsenko, Alexey and Kingma, Diederik P and Poole, Ben and Norouzi, Mohammad and Fleet, David J and others},
  journal={arXiv preprint arXiv:2210.02303},
  year={2022}
}

@article{singer2022make,
  title={Make-{A}-{V}ideo: {T}ext-to-video generation without text-video data},
  author={Singer, Uriel and Polyak, Adam and Hayes, Thomas and Yin, Xi and An, Jie and Zhang, Songyang and Hu, Qiyuan and Yang, Harry and Ashual, Oron and Gafni, Oran and others},
  journal={arXiv preprint arXiv:2209.14792},
  year={2022}
}

@article{blattmann2023stable,
  title={Stable video diffusion: {S}caling latent video diffusion models to large datasets},
  author={Blattmann, Andreas and Dockhorn, Tim and Kulal, Sumith and Mendelevitch, Daniel and Kilian, Maciej and Lorenz, Dominik and Levi, Yam and English, Zion and Voleti, Vikram and Letts, Adam and others},
  journal={arXiv preprint arXiv:2311.15127},
  year={2023}
}

@inproceedings{golinski2020feedback,
  title={Feedback recurrent autoencoder for video compression},
  author={Golinski, Adam and Pourreza, Reza and Yang, Yang and Sautiere, Guillaume and Cohen, Taco S},
  booktitle={Proceedings of the Asian Conference on Computer Vision},
  year={2020}
}

@article{d2017autoencoder,
  title={Autoencoder with recurrent neural networks for video forgery detection},
  author={D'Avino, Dario and Cozzolino, Davide and Poggi, Giovanni and Verdoliva, Luisa},
  journal={arXiv preprint arXiv:1708.08754},
  year={2017}
}

@article{zhou2024allegro,
  title={Allegro: {O}pen the black box of commercial-level video generation model},
  author={Zhou, Yuan and Wang, Qiuyue and Cai, Yuxuan and Yang, Huan},
  journal={arXiv preprint arXiv:2410.15458},
  year={2024}
}

@article{zheng2024open,
  title={Open-{S}ora: {D}emocratizing efficient video production for all},
  author={Zheng, Zangwei and Peng, Xiangyu and Yang, Tianji and Shen, Chenhui and Li, Shenggui and Liu, Hongxin and Zhou, Yukun and Li, Tianyi and You, Yang},
  journal={arXiv preprint arXiv:2412.20404},
  year={2024}
}

@article{yu2023language,
  title={Language Model Beats Diffusion--{T}okenizer is Key to Visual Generation},
  author={Yu, Lijun and Lezama, Jos{\'e} and Gundavarapu, Nitesh B and Versari, Luca and Sohn, Kihyuk and Minnen, David and Cheng, Yong and Birodkar, Vighnesh and Gupta, Agrim and Gu, Xiuye and others},
  journal={arXiv preprint arXiv:2310.05737},
  year={2023}
}

@inproceedings{pascanu2013difficulty,
  title={On the difficulty of training recurrent neural networks},
  author={Pascanu, Razvan and Mikolov, Tomas and Bengio, Yoshua},
  booktitle={International conference on machine learning},
  pages={1310--1318},
  year={2013},
  organization={Pmlr}
}

@incollection{williams2013gradient,
  title={Gradient-based learning algorithms for recurrent networks and their computational complexity},
  author={Williams, Ronald J and Zipser, David},
  booktitle={Backpropagation},
  pages={433--486},
  year={2013},
  publisher={Psychology Press}
}

@article{salehinejad2017recent,
  title={Recent advances in recurrent neural networks},
  author={Salehinejad, Hojjat and Sankar, Sharan and Barfett, Joseph and Colak, Errol and Valaee, Shahrokh},
  journal={arXiv preprint arXiv:1801.01078},
  year={2017}
}

@techreport{rumelhart1985learning,
  title={Learning internal representations by error propagation},
  author={Rumelhart, David E and Hinton, Geoffrey E and Williams, Ronald J},
  year={1985},
  institution={Institute of Cognitive Science}
}

@article{xing2024survey,
  title={A survey on video diffusion models},
  author={Xing, Zhen and Feng, Qijun and Chen, Haoran and Dai, Qi and Hu, Han and Xu, Hang and Wu, Zuxuan and Jiang, Yu-Gang},
  journal={ACM Computing Surveys},
  volume={57},
  number={2},
  pages={1--42},
  year={2024},
  publisher={ACM New York, NY}
}

@inproceedings{ceylan2023pix2video,
  title={Pix2{V}ideo: {V}ideo editing using image diffusion},
  author={Ceylan, Duygu and Huang, Chun-Hao P and Mitra, Niloy J},
  booktitle={Proceedings of the IEEE/CVF International Conference on Computer Vision},
  pages={23206--23217},
  year={2023}
}

@inproceedings{chen2024videocrafter2,
  title={Video{C}rafter2: {O}vercoming data limitations for high-quality video diffusion models},
  author={Chen, Haoxin and Zhang, Yong and Cun, Xiaodong and Xia, Menghan and Wang, Xintao and Weng, Chao and Shan, Ying},
  booktitle={Proceedings of the IEEE/CVF Conference on Computer Vision and Pattern Recognition},
  pages={7310--7320},
  year={2024}
}

@article{melnik2024video,
  title={Video diffusion models: {A} survey},
  author={Melnik, Andrew and Ljubljanac, Michal and Lu, Cong and Yan, Qi and Ren, Weiming and Ritter, Helge},
  journal={arXiv preprint arXiv:2405.03150},
  year={2024}
}

@inproceedings{blattmann2023align,
  title={Align {Y}our {L}atents: {H}igh-resolution video synthesis with latent diffusion models},
  author={Blattmann, Andreas and Rombach, Robin and Ling, Huan and Dockhorn, Tim and Kim, Seung Wook and Fidler, Sanja and Kreis, Karsten},
  booktitle={Proceedings of the IEEE/CVF conference on computer vision and pattern recognition},
  pages={22563--22575},
  year={2023}
}

@inproceedings{yu2023video,
  title={Video probabilistic diffusion models in projected latent space},
  author={Yu, Sihyun and Sohn, Kihyuk and Kim, Subin and Shin, Jinwoo},
  booktitle={Proceedings of the IEEE/CVF conference on computer vision and pattern recognition},
  pages={18456--18466},
  year={2023}
}

@article{he2022latent,
  title={Latent video diffusion models for high-fidelity long video generation},
  author={He, Yingqing and Yang, Tianyu and Zhang, Yong and Shan, Ying and Chen, Qifeng},
  journal={arXiv preprint arXiv:2211.13221},
  year={2022}
}

@article{hacohen2024ltx,
  title={{LTX}-{V}ideo: {R}ealtime video latent diffusion},
  author={HaCohen, Yoav and Chiprut, Nisan and Brazowski, Benny and Shalem, Daniel and Moshe, Dudu and Richardson, Eitan and Levin, Eran and Shiran, Guy and Zabari, Nir and Gordon, Ori and others},
  journal={arXiv preprint arXiv:2501.00103},
  year={2024}
}

@article{wang2025lavie,
  title={{LAVIE}: {H}igh-quality video generation with cascaded latent diffusion models},
  author={Wang, Yaohui and Chen, Xinyuan and Ma, Xin and Zhou, Shangchen and Huang, Ziqi and Wang, Yi and Yang, Ceyuan and He, Yinan and Yu, Jiashuo and Yang, Peiqing and others},
  journal={International Journal of Computer Vision},
  volume={133},
  number={5},
  pages={3059--3078},
  year={2025},
  publisher={Springer}
}

@article{an2023latent,
  title={Latent-{S}hift: {L}atent diffusion with temporal shift for efficient text-to-video generation},
  author={An, Jie and Zhang, Songyang and Yang, Harry and Gupta, Sonal and Huang, Jia-Bin and Luo, Jiebo and Yin, Xi},
  journal={arXiv preprint arXiv:2304.08477},
  year={2023}
}

@inproceedings{danier2024ldmvfi,
  title={{LDMVFI}: {V}ideo frame interpolation with latent diffusion models},
  author={Danier, Duolikun and Zhang, Fan and Bull, David},
  booktitle={Proceedings of the AAAI Conference on Artificial Intelligence},
  volume={38},
  pages={1472--1480},
  year={2024}
}

@inproceedings{li2024drivingdiffusion,
  title={Driving{D}iffusion: {L}ayout-guided multi-view driving scenarios video generation with latent diffusion model},
  author={Li, Xiaofan and Zhang, Yifu and Ye, Xiaoqing},
  booktitle={European Conference on Computer Vision},
  pages={469--485},
  year={2024},
  organization={Springer}
}

@article{chen2024od,
  title={{OD-VAE}: {A}n omni-dimensional video compressor for improving latent video diffusion model},
  author={Chen, Liuhan and Li, Zongjian and Lin, Bin and Zhu, Bin and Wang, Qian and Yuan, Shenghai and Zhou, Xing and Cheng, Xinhua and Yuan, Li},
  journal={arXiv preprint arXiv:2409.01199},
  year={2024}
}

@article{gao2024ca2,
  title={Ca2-{VDM}: {E}fficient autoregressive video diffusion model with causal generation and cache sharing},
  author={Gao, Kaifeng and Shi, Jiaxin and Zhang, Hanwang and Wang, Chunping and Xiao, Jun and Chen, Long},
  journal={arXiv preprint arXiv:2411.16375},
  year={2024}
}

@inproceedings{zhou2024upscale,
  title={Upscale-{A}-{V}ideo: {T}emporal-consistent diffusion model for real-world video super-resolution},
  author={Zhou, Shangchen and Yang, Peiqing and Wang, Jianyi and Luo, Yihang and Loy, Chen Change},
  booktitle={Proceedings of the IEEE/CVF Conference on Computer Vision and Pattern Recognition},
  pages={2535--2545},
  year={2024}
}

@article{xie2025star,
  title={{STAR}: {S}patial-temporal augmentation with text-to-video models for real-world video super-resolution},
  author={Xie, Rui and Liu, Yinhong and Zhou, Penghao and Zhao, Chen and Zhou, Jun and Zhang, Kai and Zhang, Zhenyu and Yang, Jian and Yang, Zhenheng and Tai, Ying},
  journal={arXiv preprint arXiv:2501.02976},
  year={2025}
}

@inproceedings{chen2025dove,
    title={{DOVE}: {E}fficient One-Step Diffusion Model for Real-World Video Super-Resolution},
    author={Zheng Chen and Zichen Zou and Kewei Zhang and Xiongfei Su and Xin Yuan and Yong Guo and Yulun Zhang},
    booktitle={The Thirty-ninth Annual Conference on Neural Information Processing Systems},
    year={2025},
}

@article{chen2024mvsplat360,
  title={{MVS}plat360: {F}eed-forward 360 scene synthesis from sparse views},
  author={Chen, Yuedong and Zheng, Chuanxia and Xu, Haofei and Zhuang, Bohan and Vedaldi, Andrea and Cham, Tat-Jen and Cai, Jianfei},
  journal={Advances in Neural Information Processing Systems},
  volume={37},
  pages={107064--107086},
  year={2024}
}

@article{ding2020iqa,
  title={Image Quality Assessment: {U}nifying Structure and Texture Similarity},
  author={Ding, Keyan and Ma, Kede and Wang, Shiqi and Simoncelli, Eero P.},
  journal = {CoRR},
  volume = {abs/2004.07728},
  year={2020},
  url = {https://arxiv.org/abs/2004.07728}
}

@inproceedings{zhang2018unreasonable,
  title={The unreasonable effectiveness of deep features as a perceptual metric},
  author={Zhang, Richard and Isola, Phillip and Efros, Alexei A and Shechtman, Eli and Wang, Oliver},
  booktitle={Proceedings of the IEEE conference on computer vision and pattern recognition},
  pages={586--595},
  year={2018}
}

@inproceedings{wang2021real,
  title={Real-{ESRGAN}: {T}raining real-world blind super-resolution with pure synthetic data},
  author={Wang, Xintao and Xie, Liangbin and Dong, Chao and Shan, Ying},
  booktitle={Proceedings of the IEEE/CVF international conference on computer vision},
  pages={1905--1914},
  year={2021}
}

@article{yue2023resshift,
  title={Res{S}hift: {E}fficient diffusion model for image super-resolution by residual shifting},
  author={Yue, Zongsheng and Wang, Jianyi and Loy, Chen Change},
  journal={Advances in Neural Information Processing Systems},
  volume={36},
  pages={13294--13307},
  year={2023}
}

@inproceedings{chan2022investigating,
  title={Investigating tradeoffs in real-world video super-resolution},
  author={Chan, Kelvin CK and Zhou, Shangchen and Xu, Xiangyu and Loy, Chen Change},
  booktitle={Proceedings of the IEEE/CVF conference on computer vision and pattern recognition},
  pages={5962--5971},
  year={2022}
}

@inproceedings{yang2024motion,
  title={Motion-{G}uided latent diffusion for temporally consistent real-world video super-resolution},
  author={Yang, Xi and He, Chenhang and Ma, Jianqi and Zhang, Lei},
  booktitle={European conference on computer vision},
  pages={224--242},
  year={2024},
  organization={Springer}
}

@article{he2024venhancer,
  title={V{E}nhancer: {G}enerative space-time enhancement for video generation},
  author={He, Jingwen and Xue, Tianfan and Liu, Dongyang and Lin, Xinqi and Gao, Peng and Lin, Dahua and Qiao, Yu and Ouyang, Wanli and Liu, Ziwei},
  journal={arXiv preprint arXiv:2407.07667},
  year={2024}
}

@inproceedings{ling2024dl3dv,
  title={{DL3DV}-10k: {A} large-scale scene dataset for deep learning-based 3{D} vision},
  author={Ling, Lu and Sheng, Yichen and Tu, Zhi and Zhao, Wentian and Xin, Cheng and Wan, Kun and Yu, Lantao and Guo, Qianyu and Yu, Zixun and Lu, Yawen and others},
  booktitle={Proceedings of the IEEE/CVF Conference on Computer Vision and Pattern Recognition},
  year={2024}
}

@inproceedings{wu2025difix3d,
  title={Difix3{D}+: {I}mproving 3{D} reconstructions with single-step diffusion models},
  author={Wu, Jay Zhangjie and Zhang, Yuxuan and Turki, Haithem and Ren, Xuanchi and Gao, Jun and Shou, Mike Zheng and Fidler, Sanja and Gojcic, Zan and Ling, Huan},
  booktitle={Proceedings of the Computer Vision and Pattern Recognition Conference},
  year={2025}
}

@article{mildenhall2021nerf,
  title={Nerf: Representing scenes as neural radiance fields for view synthesis},
  author={Mildenhall, Ben and Srinivasan, Pratul P and Tancik, Matthew and Barron, Jonathan T and Ramamoorthi, Ravi and Ng, Ren},
  journal={Communications of the ACM},
  volume={65},
  number={1},
  year={2021},
  publisher={ACM New York, NY, USA}
}

@article{kerbl20233d,
  title={3D Gaussian splatting for real-time radiance field rendering.},
  author={Kerbl, Bernhard and Kopanas, Georgios and Leimk{\"u}hler, Thomas and Drettakis, George},
  journal={ACM Trans. Graph.},
  volume={42},
  number={4},
  year={2023}
}

@article{ye2025gsplat,
  title={gsplat: An open-source library for Gaussian splatting},
  author={Ye, Vickie and Li, Ruilong and Kerr, Justin and Turkulainen, Matias and Yi, Brent and Pan, Zhuoyang and Seiskari, Otto and Ye, Jianbo and Hu, Jeffrey and Tancik, Matthew and others},
  journal={Journal of Machine Learning Research},
  volume={26},
  number={34},
  year={2025}
}

@article{zhao2023pytorch,
  title={Py{T}orch {FSDP}: {E}xperiences on scaling fully sharded data parallel},
  author={Zhao, Yanli and Gu, Andrew and Varma, Rohan and Luo, Liang and Huang, Chien-Chin and Xu, Min and Wright, Less and Shojanazeri, Hamid and Ott, Myle and Shleifer, Sam and others},
  journal={arXiv preprint arXiv:2304.11277},
  year={2023}
}

@inproceedings{sauer2024adversarial,
  title={Adversarial diffusion distillation},
  author={Sauer, Axel and Lorenz, Dominik and Blattmann, Andreas and Rombach, Robin},
  booktitle={European Conference on Computer Vision},
  year={2024},
  organization={Springer}
}

@article{yi2019multi,
  title={Multi-temporal ultra dense memory network for video super-resolution},
  author={Yi, Peng and Wang, Zhongyuan and Jiang, Kui and Shao, Zhenfeng and Ma, Jiayi},
  journal={IEEE Transactions on Circuits and Systems for Video Technology},
  volume={30},
  number={8},
  year={2019},
  publisher={IEEE}
}

@InProceedings{tao2017spmc,
  author    = {Xin Tao and
               Hongyun Gao and
               Renjie Liao and
               Jue Wang and
               Jiaya Jia},
  title = {Detail-Revealing Deep Video Super-Resolution},
  booktitle = {The IEEE International Conference on Computer Vision (ICCV)},
  month = {Oct},
  year = {2017}
}

@inproceedings{zhou2024upscaleavideo,
   title={{Upscale-A-Video}: {T}emporal-Consistent Diffusion Model for Real-World Video Super-Resolution},
   author={Zhou, Shangchen and Yang, Peiqing and Wang, Jianyi and Luo, Yihang and Loy, Chen Change},
   booktitle={CVPR},
   year={2024}
}

@article{yang2021real,
  title={Real-world Video Super-resolution: {A} Benchmark Dataset and A Decomposition based Learning Scheme},
  author={YANG, Xi and Xiang, Wangmeng and Zeng, Hui and Zhang, Lei},
  journal={ICCV},
  year={2021}
}

@inproceedings{wang2023benchmark,
  title={Benchmark dataset and effective inter-frame alignment for real-world video super-resolution},
  author={Wang, Ruohao and Liu, Xiaohui and Zhang, Zhilu and Wu, Xiaohe and Feng, Chun-Mei and Zhang, Lei and Zuo, Wangmeng},
  booktitle={Proceedings of the IEEE/CVF conference on computer vision and pattern recognition},
  year={2023}
}

@inproceedings{jiang2025vace,
  title={Vace: All-in-one video creation and editing},
  author={Jiang, Zeyinzi and Han, Zhen and Mao, Chaojie and Zhang, Jingfeng and Pan, Yulin and Liu, Yu},
  booktitle={Proceedings of the IEEE/CVF International Conference on Computer Vision},
  pages={17191--17202},
  year={2025}
}

@inproceedings{waymodataset,
  title={Scalability in perception for autonomous driving: Waymo open dataset},
  author={Sun, Pei and Kretzschmar, Henrik and Dotiwalla, Xerxes and Chouard, Aurelien and Patnaik, Vijaysai and Tsui, Paul and Guo, James and Zhou, Yin and Chai, Yuning and Caine, Benjamin and others},
  booktitle={Proceedings of the IEEE/CVF conference on computer vision and pattern recognition},
  pages={2446--2454},
  year={2020}
}

@inproceedings{rovi,
  title={Towards language-driven video inpainting via multimodal large language models},
  author={Wu, Jianzong and Li, Xiangtai and Si, Chenyang and Zhou, Shangchen and Yang, Jingkang and Zhang, Jiangning and Li, Yining and Chen, Kai and Tong, Yunhai and Liu, Ziwei and others},
  booktitle={Proceedings of the IEEE/CVF Conference on Computer Vision and Pattern Recognition},
  pages={12501--12511},
  year={2024}
}

@article{wang2025mirage,
  title={Mirage: One-Step Video Diffusion for Photorealistic and Coherent Asset Editing in Driving Scenes},
  author={Wang, Shuyun and Sun, Haiyang and Wang, Bing and Ye, Hangjun and Yu, Xin},
  journal={arXiv preprint arXiv:2512.24227},
  year={2025}
}

@article{ljungbergh2025r3d2,
  title={R3d2: Realistic 3d asset insertion via diffusion for autonomous driving simulation},
  author={Ljungbergh, William and Taveira, Bernardo and Zheng, Wenzhao and Tonderski, Adam and Peng, Chensheng and Kahl, Fredrik and Petersson, Christoffer and Felsberg, Michael and Keutzer, Kurt and Tomizuka, Masayoshi and others},
  journal={arXiv preprint arXiv:2506.07826},
  year={2025}
}

@article{pix2pixturbo,
  title={One-step image translation with text-to-image models},
  author={Parmar, Gaurav and Park, Taesung and Narasimhan, Srinivasa and Zhu, Jun-Yan},
  journal={arXiv preprint arXiv:2403.12036},
  year={2024}
}

@article{dmd,
  title={Improved distribution matching distillation for fast image synthesis},
  author={Yin, Tianwei and Gharbi, Micha{\"e}l and Park, Taesung and Zhang, Richard and Shechtman, Eli and Durand, Fredo and Freeman, Bill},
  journal={Advances in neural information processing systems},
  volume={37},
  pages={47455--47487},
  year={2024}
}

@inproceedings{hess2025splatad,
  title={Splatad: Real-time lidar and camera rendering with 3d gaussian splatting for autonomous driving},
  author={Hess, Georg and Lindstr{\"o}m, Carl and Fatemi, Maryam and Petersson, Christoffer and Svensson, Lennart},
  booktitle={Proceedings of the Computer Vision and Pattern Recognition Conference},
  pages={11982--11992},
  year={2025}
}

@inproceedings{trellis,
  title={Structured 3d latents for scalable and versatile 3d generation},
  author={Xiang, Jianfeng and Lv, Zelong and Xu, Sicheng and Deng, Yu and Wang, Ruicheng and Zhang, Bowen and Chen, Dong and Tong, Xin and Yang, Jiaolong},
  booktitle={Proceedings of the IEEE/CVF conference on computer vision and pattern recognition},
  pages={21469--21480},
  year={2025}
}

@article{orient_anything,
  title={Orient anything: Learning robust object orientation estimation from rendering 3d models},
  author={Wang, Zehan and Zhang, Ziang and Pang, Tianyu and Du, Chao and Zhao, Hengshuang and Zhao, Zhou},
  journal={arXiv preprint arXiv:2412.18605},
  year={2024}
}

@inproceedings{sdturbo,
  title={Fast high-resolution image synthesis with latent adversarial diffusion distillation},
  author={Sauer, Axel and Boesel, Frederic and Dockhorn, Tim and Blattmann, Andreas and Esser, Patrick and Rombach, Robin},
  booktitle={SIGGRAPH Asia 2024 Conference Papers},
  pages={1--11},
  year={2024}
}

@article{self-forcing,
  title={Self forcing: Bridging the train-test gap in autoregressive video diffusion},
  author={Huang, Xun and Li, Zhengqi and He, Guande and Zhou, Mingyuan and Shechtman, Eli},
  journal={arXiv preprint arXiv:2506.08009},
  year={2025}
}

@inproceedings{ljungbergh2024neuroncap,
  title={Neuroncap: Photorealistic closed-loop safety testing for autonomous driving},
  author={Ljungbergh, William and Tonderski, Adam and Johnander, Joakim and Caesar, Holger and {\AA}str{\"o}m, Kalle and Felsberg, Michael and Petersson, Christoffer},
  booktitle={European Conference on Computer Vision},
  pages={161--177},
  year={2024},
  organization={Springer}
}

@inproceedings{ost2021neural,
  title={Neural scene graphs for dynamic scenes},
  author={Ost, Julian and Mannan, Fahim and Thuerey, Nils and Knodt, Julian and Heide, Felix},
  booktitle={Proceedings of the IEEE/CVF Conference on Computer Vision and Pattern Recognition},
  pages={2856--2865},
  year={2021}
}

@article{zhou2025hugsim,
  title={Hugsim: A real-time, photo-realistic and closed-loop simulator for autonomous driving},
  author={Zhou, Hongyu and Lin, Longzhong and Wang, Jiabao and Lu, Yichong and Bai, Dongfeng and Liu, Bingbing and Wang, Yue and Geiger, Andreas and Liao, Yiyi},
  journal={IEEE Transactions on Pattern Analysis and Machine Intelligence},
  year={2025},
  publisher={IEEE}
}

@article{yang2025towards,
  title={Towards One-step Causal Video Generation via Adversarial Self-Distillation},
  author={Yang, Yongqi and Huang, Huayang and Peng, Xu and Hu, Xiaobin and Luo, Donghao and Zhang, Jiangning and Wang, Chengjie and Wu, Yu},
  journal={arXiv preprint arXiv:2511.01419},
  year={2025}
}

@inproceedings{wang2025videoscene,
  title={Videoscene: Distilling video diffusion model to generate 3d scenes in one step},
  author={Wang, Hanyang and Liu, Fangfu and Chi, Jiawei and Duan, Yueqi},
  booktitle={2025 IEEE/CVF Conference on Computer Vision and Pattern Recognition (CVPR)},
  pages={16475--16485},
  year={2025},
  organization={IEEE}
}

@inproceedings{noroozi2024you,
  title={You only need one step: Fast super-resolution with stable diffusion via scale distillation},
  author={Noroozi, Mehdi and Hadji, Isma and Martinez, Brais and Bulat, Adrian and Tzimiropoulos, Georgios},
  booktitle={European Conference on Computer Vision},
  pages={145--161},
  year={2024},
  organization={Springer}
}

@inproceedings{mao2025osv,
  title={Osv: One step is enough for high-quality image to video generation},
  author={Mao, Xiaofeng and Jiang, Zhengkai and Wang, Fu-Yun and Zhang, Jiangning and Chen, Hao and Chi, Mingmin and Wang, Yabiao and Luo, Wenhan},
  booktitle={Proceedings of the Computer Vision and Pattern Recognition Conference},
  pages={12585--12594},
  year={2025}
}

@article{he2024one,
  title={One step diffusion-based super-resolution with time-aware distillation},
  author={He, Xiao and Tang, Huaao and Tu, Zhijun and Zhang, Junchao and Cheng, Kun and Chen, Hanting and Guo, Yong and Zhu, Mingrui and Wang, Nannan and Gao, Xinbo and others},
  journal={arXiv preprint arXiv:2408.07476},
  year={2024}
}

@inproceedings{wang2024sinsr,
  title={Sinsr: diffusion-based image super-resolution in a single step},
  author={Wang, Yufei and Yang, Wenhan and Chen, Xinyuan and Wang, Yaohui and Guo, Lanqing and Chau, Lap-Pui and Liu, Ziwei and Qiao, Yu and Kot, Alex C and Wen, Bihan},
  booktitle={Proceedings of the IEEE/CVF conference on computer vision and pattern recognition},
  pages={25796--25805},
  year={2024}
}

@article{wang2025seedvr2,
  title={Seedvr2: One-step video restoration via diffusion adversarial post-training},
  author={Wang, Jianyi and Lin, Shanchuan and Lin, Zhijie and Ren, Yuxi and Wei, Meng and Yue, Zongsheng and Zhou, Shangchen and Chen, Hao and Zhao, Yang and Yang, Ceyuan and others},
  journal={arXiv preprint arXiv:2506.05301},
  year={2025}
}

@article{dong2026one,
  title={One-Shot Refiner: Boosting Feed-forward Novel View Synthesis via One-Step Diffusion},
  author={Dong, Yitong and Zhang, Qi and Jiang, Minchao and Wu, Zhiqiang and Fan, Qingnan and Feng, Ying and Zhang, Huaqi and Bao, Hujun and Zhang, Guofeng},
  journal={arXiv preprint arXiv:2601.14161},
  year={2026}
}

@article{teng2025gfix,
  title={GFix: Perceptually Enhanced Gaussian Splatting Video Compression},
  author={Teng, Siyue and Gao, Ge and Danier, Duolikun and Jiang, Yuxuan and Zhang, Fan and Davis, Thomas and Liu, Zoe and Bull, David},
  journal={arXiv preprint arXiv:2511.06953},
  year={2025}
}

@inproceedings{chadebec2025flash,
  title={Flash diffusion: Accelerating any conditional diffusion model for few steps image generation},
  author={Chadebec, Clement and Tasar, Onur and Benaroche, Eyal and Aubin, Benjamin},
  booktitle={Proceedings of the AAAI Conference on Artificial Intelligence},
  volume={39},
  pages={15686--15695},
  year={2025}
}

@article{lee2025single,
  title={Single-Step Bidirectional Unpaired Image Translation Using Implicit Bridge Consistency Distillation},
  author={Lee, Suhyeon and Kim, Kwanyoung and Ye, Jong Chul},
  journal={arXiv preprint arXiv:2503.15056},
  year={2025}
}

@inproceedings{chen2025genhaze,
  title={GenHaze: Pioneering Controllable One-Step Realistic Haze Generation for Real-World Dehazing},
  author={Chen, Sixiang and Ye, Tian and Lin, Yunlong and Jin, Yeying and Yang, Yijun and Chen, Haoyu and Lai, Jianyu and Fei, Song and Xing, Zhaohu and Tsung, Fugee and others},
  booktitle={Proceedings of the IEEE/CVF International Conference on Computer Vision},
  pages={9194--9205},
  year={2025}
}

@InProceedings{huang2023vbench,
      title={{VBench}: Comprehensive Benchmark Suite for Video Generative Models},
      author={Huang, Ziqi and He, Yinan and Yu, Jiashuo and Zhang, Fan and Si, Chenyang and Jiang, Yuming and Zhang, Yuanhan and Wu, Tianxing and Jin, Qingyang and Chanpaisit, Nattapol and Wang, Yaohui and Chen, Xinyuan and Wang, Limin and Lin, Dahua and Qiao, Yu and Liu, Ziwei},
      booktitle={Proceedings of the IEEE/CVF Conference on Computer Vision and Pattern Recognition},
      year={2024}
}
